\documentclass[letterpaper, 10 pt, conference]{ieeeconf}  

\usepackage{amsmath}
\usepackage{amssymb} 

\renewcommand{\QED}{\QEDopen}

\usepackage[labelformat=simple]{subcaption}
\DeclareCaptionLabelSeparator{periodspace}{.\quad}
\usepackage{tikz}
\usetikzlibrary{arrows,matrix,positioning,patterns}
\usetikzlibrary{calc}
\usepackage{pgfplots} 
\usetikzlibrary{plotmarks}

\usepackage[utf8]{inputenc}
\usepackage{pgfplots}
\usepgfplotslibrary{groupplots}

\usepackage{url}
\usepackage{multirow}
\usepackage{colortbl}

\usepackage[ruled, vlined, linesnumbered]{algorithm2e}

\DeclareSymbolFont{bbold}{U}{bbold}{m}{n}
\DeclareSymbolFontAlphabet{\mathbbold}{bbold}

\usepackage[labelformat=simple]{subcaption}
\DeclareCaptionLabelSeparator{periodspace}{.\quad}
\usepackage{listings}
\usepackage{xcolor}
\usepackage{cite}

\usepackage{wrapfig}
\usepackage{supertabular}

\IEEEoverridecommandlockouts                              
\title{\LARGE \bf
Online Sampling in the Parameter Space of a Neural Network for GPU-accelerated Motion Planning of Autonomous Vehicles}

\author{Mogens Graf Plessen
\thanks{MPG is an independent researcher, {\tt\small mgplessen@gmail.com}}
}

\renewcommand{\baselinestretch}{0.98}
\begin{document}

\maketitle
\thispagestyle{empty}
\pagestyle{empty}

%
%
%


\begin{abstract}
This paper proposes online sampling in the parameter space of a neural network for GPU-accelerated motion planning of autonomous vehicles. Neural networks are used as controller parametrization since they can handle nonlinear non-convex systems and their complexity does not scale with prediction horizon length. Network parametrizations are sampled at each sampling time and then held constant throughout the prediction horizon. Controls still vary over the prediction horizon due to varying feature vectors fed to the network. Full-dimensional vehicles are modeled by  polytopes. Under the assumption of obstacle point data, and their extrapolation over a prediction horizon under constant velocity assumption, collision avoidance reduces to linear inequality checks. Steering and longitudinal acceleration controls are determined simultaneously. The proposed method is designed for parallelization and therefore well-suited to benefit from continuing advancements in hardware such as GPUs. Characteristics of proposed method are illustrated in 5 numerical simulation experiments including dynamic obstacle avoidance, waypoint tracking requiring alternating forward and reverse driving with maximal steering, and a reverse parking scenario.
\end{abstract}

\section{Introduction and Problem Formulation\label{sec_intro}}

\subsection{Motivation\label{subsec_introMotiv}}

This paper is motivated by the desire for a simple control scheme that can (i) be based on arbitrarily complex nonlinear non-convex vehicle models, (ii) work for general all-purpose trajectory planning  (i.e., is equally applicable to scenarios from parking to road centerline tracking), (iii) can generate collision-free trajectories accounting for the full vehicle dimensions, and (iv) can  exploit continuing advancement in computation hardware, in particular, for parallelization.

\subsection{Problem formulation and contribution\label{subsec_problFormulation}}

\begin{figure}
\centering
\begin{tikzpicture}
\node[color=black] (a) at (-1, 0.88) { \small{human route} };
\node[color=black] (a) at (-1.23, 0.57) { \small{selection} };
\draw[->,>=latex'] (-0.05, 0.72) -- (0.75,0.72);
\draw[fill=white,solid] (0.75,0.45) rectangle (1.65,0.95);
\node[color=black] (a) at (1.2, 0.72) { Navi };
\draw[->,>=latex'] (1.2, 0.45) -- (1.2,0.15);
\draw[fill=white,solid] (0.75,-0.45) rectangle (1.65,0.15);
\node[color=black] (a) at (1.2, -0.15) { Filter };
\draw[->,>=latex'] (1.65, -0.15) -- (2.65,-0.15);
\node[color=black] (a) at (2.12, 0.03) { \small{\textbf{$s_t$}}};
\draw[fill=white,thick] (2.65,-0.45) rectangle (3.55,0.15);
\node[color=black] (a) at (3.1, -0.15) { \textbf{C} };
\node[color=black] (a) at (4, 0) { \small{\textbf{$a_t$}}};
\draw[->,>=latex'] (3.55, -0.15) -- (4.4,-0.15);
\draw[fill=white,solid] (4.4,-0.5) rectangle (5.4,0.2);
\draw[fill=white,solid] (4.6,-0.3) rectangle (5.2,-0.15);
\draw (4.74,-0.35) circle [radius=0.05];
\draw (5.04,-0.35) circle [radius=0.05];
\draw[-] (4.65, -0.15) -- (4.68,-0.02) -- (4.9,-0.02) -- (5.066,-0.15);
\draw[-] (4.85, -0.15) -- (4.85,-0.02);
\draw[->,>=latex'] (5.4, -0.15) -- (5.9,-0.15) -- (5.9,-0.75) -- (-0.05,-0.75) -- (-0.05,0.02)--(0.75,0.02);
\draw[->,>=latex'] (-0.05, -0.32) -- (0.75,-0.32);
\node[color=black] (a) at (2.95, -1.0) { \small{extero- \& proprioceptive measurements}};
\end{tikzpicture}
\caption{Closed-loop control architecture. ``Navi'' and ``Filter'' map human route  selections as well as extero- and proprioceptive measurements to feature vector $s_t$. This paper proposes a neural network-based and GPU-accelerated online sampling algorithm for controller C, which maps $s_t$ to control action $a_t$ to be applied to the vehicle. The algorithm accounts for physical system constraints and for static and dynamic obstacle avoidance.}
\label{fig:closed_loop}
\vspace{-0.3cm}
\end{figure}
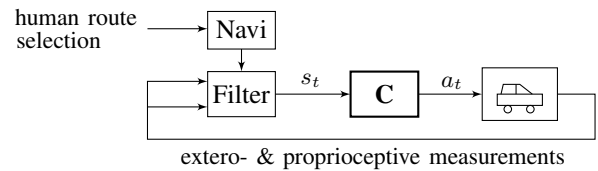

The problem addressed is to design a method to generate control signals (e.g., steering and longitudinal acceleration) such that a full-dimensional vehicle can drive automatedly from an initial vehicle state to a desired goal state while avoiding static and dynamic obstacles, and accounting for sensor measurements (perception) and typical environmental constraints (traffic rules). Such a method is visualized abstractly in Fig. \ref{fig:closed_loop}. This paper focuses on the control aspect, i.e., the design of C. Thus, it is assumed throughout that general high-level route, obstacle points data and other relevant environment measurements are made available by a navigation and perception module not subject of this paper.

The contribution of this paper is a control method that parameterizes controller C in Fig. \ref{fig:closed_loop} by a neural network and and employs a specific GPU-accelerated gradient-free algorithm for the online sampling of its parameters every sampling time such that all initial motivating aspects (i)-(iv) from Sect. \ref{subsec_introMotiv} are addressed. 
 
\subsection{Background and further motivation}

In the survey of \cite{gonzalez2016review} motion planning techniques for automated vehicles are classified into 4 groups: graph-search, sampling, interpolating curve and optimal control based planners. The typical ingredients common to all 4 are: (i) a mathematical vehicle model (possibly nonlinear, non-holonomic and non-convex), (ii) a priori mission information (at least start and goal location; often also high-level map-based route information such as road-centerline references, velocities, etc.), (iii) an exteroceptive and a proprioceptive sensors setup, and (iv) a vehicle's actuators setup. Then, the 4 groups differ in how they map information (i)-(iii) to control signals fed to (iv) such that the resulting vehicle motion is collision-free. This problem is complex \cite{reif1979complexity}. Therefore, all motion planning techniques typically make additional specific assumptions for their algorithms to work. For example, graph-search and optimal control methods often assume specific obstacle shapes\cite{lozano1979algorithm,brown2017safe,andersen2017trajectory,
plessen2018spatial,oyama2013model}, which necessitates first an upstream mapping of raw sensor data to such obstacle descriptions. In \cite{li2019real,hamer2019fast} ground robots are approximated as circles to simplify collision checks. Vehicle models employed are of great variety, (i) often differing for tasks (parking, tube-like road driving, limits of handling, etc.), (ii) expressed in different coordinate systems (absolute or road aligned), and (iii) even varying over different hierarchies, for example, when differentiating trajectory generation and consequent tracking \cite{berntorp2017path}. For (i)-(ii), switching logics and multiple controller designs are required. For (iii), hierarchy-encroaching feasibility issues may be encountered.

Sampling based motion planners or, in general, methods that use randomization are attractive since these are \emph{probabilistic complete}: a solution will be found with probability 1 when the simulation effort goes to infinity and a solution exists. Many of such algorithms are founded on RRTs~\cite{lavalle1998rapidly} because of its characteristic expansion of the transition tree heavily biased toward unexplored space. However, randomization comes at a cost. For RRTs, jagged controls, varying costs, varying number of nodes, and varying solution times are typical. This resulted in a large body of work improving the basic RRT-algorithm via heuristics for specific applications \cite{kuffner2000rrt, urmson2003approaches, ferguson2006replanning, kuwata2009real, bialkowski2011massively}. One issue are ``jagged'' paths as reported already in the original RRT-paper \cite{lavalle1998rapidly}, but also in alternatively popular \emph{search-based} methods~\cite{dolgov2010path}. In the latter, a post-processing smoothing step is employed involving both nonlinear optimization plus then interpolation.

The proposed algorithm belongs to the group of sampling based planners. However, its main characteristic is to sample online in the \emph{parameter space} of a neural network parameterizing the controller, an approach not taken in any of the references from the reviews \cite{gonzalez2016review, berntorp2017path} and \cite{paden2016survey}. 


After a summary of notation, Sect. \ref{sec_problem_soln} describes the method proposed to solve the problem formulated in Sect. \ref{subsec_problFormulation}. Numerical simulation experiments are provided in Sect. \ref{sec_expts}. More comments and limitations are summarized in Sect.~\ref{sec_discussion}, before concluding in Sect. \ref{sec_concl}.

\subsection{Notation}

\begin{tabular}{ll}
CoG & Vehicle center of gravity.\\
EV & Ego-vehicle.\\
NN & Neural network.\\
TSHC & Main algorithm from \cite{plessen2019automating}.\\
$(\cdot)_t$ & Variable at time-index $t$.\\
$n_{(\cdot)}\in\mathbb{N}_{++}$ & Variable dimension (e.g., $n_z$).\\
$N_{(\cdot)}$ & Scalar number of data points.\\
$H\in\mathbb{N}_{++}$ & Prediction horizon.\\
$\mathcal{M}\in\mathbb{R}^{N_\mathcal{M} \times n_\mathcal{M}}$ & Available a priori mission data.\\
$N_\text{restarts}\in\mathbb{N}_{++}$ & Hyperparameter in TSHC\cite{plessen2019automating}.\\
$N_\text{iter}^\text{max}\in\mathbb{N}_{++}$ & Hyperparameter in TSHC\cite{plessen2019automating}.\\
$N_\text{obstPts}\in\mathbb{N}_+$ & Nr. of obstacle points (each 4D).\\
$P^\text{min}/\bar{P}/P^\text{max}$ & Min/Avg/Max. path length.\\
$T_s\in\mathbb{R}$ & Sampling time.\\
$\theta\in\mathbb{R}^{1\times n_\theta}$ & Vector of NN-parameters.\\
$a\in\mathbb{R}^{1\times n_a}$ & Control action vector.\\
$\Delta \xi,\Delta\eta,\Delta \varphi,\Delta v$ & Normalization constants in \eqref{eq_def_stph}.\\
$\epsilon_\xi,\epsilon_\eta,\epsilon_\varphi,\epsilon_v\in\mathbb{R}$ & Tolerances in subscripted states.\\
$l_f,l_r\in\mathbb{R}$ & Distance CoG to front,rear axle.\\
$m^z\in\mathbb{R}^{1\times n_{m^z}}$ & Proprioceptive measurements.\\
$m^\text{ext}\in\mathbb{R}^{1\times n_{m^\text{ext}}}$ & Exteroceptive measurements.\\
$(m^\text{ext,x},~m^\text{ext,y})$ & $(x,~y)$-coordinates of $m^\text{ext}$.\\
$\{\hat{m}_{t+h}^\text{ext}\}_{h=0}^H$ & Extrapolation for prediction.\\
$m^\text{ref}\in\mathbb{R}^{1\times n_{m^\text{ref}}}$ & Reference measurements vector.\\
$n\in\mathbb{N}_{++}$ & Hyperparameter in TSHC\cite{plessen2019automating}.\\
$s\in\mathbb{R}^{1\times n_s}$ & Feature vector.\\
$\bar{\tau}_a\in\mathbb{R}$ & Avg. comput. time Step 5 of Alg. \ref{alg_clLoop}.\\
$t\in\mathbb{N}_+$ & Variable for indexing time:  $tT_s$.\\ 
$\varphi,~v$ & Vehicle heading angle and velocity.\\
$(x,y)$ & Vehicle CoG location.\\
$(\xi,\eta)$ & 2D EV-aligned coordinate system.\\
$z\in\mathbb{R}^{1\times n_z}$ & Vehicle state vector.\\
$\zeta^\text{goal}\in\mathbb{R}^{1\times n_{\zeta^\text{goal}}}$ & Goal setpoint vector.\\
\end{tabular}


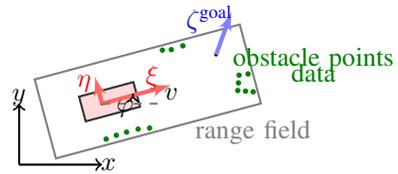
\begin{figure}
\centering
\vspace{0.2cm}
\begin{tikzpicture}[thick,scale=0.34, every node/.style={scale=1}]
\coordinate (R) at (1.8,1.55);
\coordinate (F) at ($ (R) + 4*({cos(30)},{sin(30)})$);
\coordinate (C) at ($ (R) + 1.667*({cos(30)},{sin(30)})$);
\coordinate (Ra) at ($ (C) - 2.6*({cos(30)},{sin(30)})$);
\coordinate (Raa) at ($ (C) - 2.92*({cos(30)},{sin(30)})$);
\coordinate (Rb) at ($ (C) + 3.6*({cos(30)},{sin(30)})$);
\def\psis{15} 
\def \wheelL {0.5}
\def \lineRhoSTop {2.2}
\def \lineRhoSBottom {1.7}
\draw[->] (0, 0) -- (C|-,0);
\draw[->] (0, 0) -- (0, |-C);
\node[color=black] (a) at ($ (C|-,0) + 0.25*({cos(0)},{sin(00)})$) {$x$}; 
\node[color=black] (a) at ($ (0, |-C) + 0.25*({cos(90)},{sin(90)})$) {$y$}; 
\draw[black,fill=red!90,fill opacity=0.2,rotate around={15:(C)}] ($ (C) - ({0.8},{0.5})$) rectangle ($ (C) + ({1.4},{0.5})$);
\draw[black!50,fill=white,fill opacity=0.2,rotate around={15:(C)}] ($ (C) - ({2.3},{1.6})$) rectangle ($ (C) + ({6},{1.6})$);
\fill[red] (C) circle [radius=2pt];
\draw[->,color=red!60,>=latex',ultra thick] (C) -- ($ (C) + 2.8*({cos(15)},{sin(15)})$);
\node[color=red] (a) at ($ (C) + (1.95,1.03)$) {$\xi$};
\node[color=black] (a) at ($ (C) + (2.7,0.4)$) {$v$};
\draw[->,color=red!60,>=latex',ultra thick] (C) -- ($ (C) + 1.1*({cos(105)},{sin(105)})$);
\node[color=red] (a) at ($ (C) + (-0.7,0.8)$) {$\eta$};
\draw[->, color=black] ($ (C) + 1.25*({cos(\psis)},{sin(\psis)})$) arc (\psis:15:1.25);
\node[color=black!50] (a) at ($ (C) + (5.9,-1.1)$) {range field};
\draw[color=black!50,dashed] (C) -- ($ (C) + (2.3,0)$);
\node[color=black] (a) at ($ (C) + (0.9,-0.2)$) {$\varphi$};
\fill ($ (C) + 4.8*({cos(15)},{sin(15)}) + 0.7*({cos(105)},{sin(105)})$) circle [radius=2pt];
\draw[->,color=blue!50,>=latex',ultra thick] ($ (C) + 4.8*({cos(15)},{sin(15)}) + 0.7*({cos(105)},{sin(105)})$) -- ($ (C) + 4.8*({cos(15)},{sin(15)}) + 0.7*({cos(105)},{sin(105)}) + 1.8*({cos(70)},{sin(70)}) $);
\node[color=blue] (a) at ($ (C) + (4.2,3.3)$) {$\zeta^\text{goal}$};
%
\fill[green!50!black] (8.6,3.5) circle [radius=3pt];
\fill[green!50!black] (8.55,2.9) circle [radius=3pt];
\fill[green!50!black] (9.15,2.85) circle [radius=3pt];
\fill[green!50!black] (8.85,2.875) circle [radius=3pt];
\fill[green!50!black] (8.575,3.2) circle [radius=3pt];
\fill[green!50!black] (8.95,3.55) circle [radius=3pt];
\node[color=green!50!black] (a) at (11.5,4.1) {obstacle points};
\node[color=green!50!black] (a) at (11.5,3.6) {data};
\fill[green!50!black] (3.4,1) circle [radius=3pt];
\fill[green!50!black] (3.8,1.15) circle [radius=3pt];
\fill[green!50!black] (4.2,1.27) circle [radius=3pt];
\fill[green!50!black] (4.65,1.39) circle [radius=3pt];
\fill[green!50!black] (5.1,1.43) circle [radius=3pt];
\fill[green!50!black] (5.95,4.5) circle [radius=3pt];
\fill[green!50!black] (6.4,4.65) circle [radius=3pt];
\fill[green!50!black] (5.6,4.48) circle [radius=3pt];
\end{tikzpicture}
\caption{Visualization of $x$, $y$, $\varphi$ and $v$, vehicle-aligned coordinates $\xi$ and $\eta$, an exemplary setpoint $\zeta^\text{goal}$ and measurable obstacle points data within a range field surrounding an, e.g., rectangular vehicle (red) in the 2D plane.}
\label{fig_2Dsketch}
\vspace{-0.2cm}
\end{figure}


\newlength\figureheight
\newlength\figurewidth 
\setlength\figureheight{6.5cm}
\setlength\figurewidth{6.5cm}
\begin{figure*}
\begin{tabular}{p{0.15\textwidth} p{0.57\textwidth} p{0.12\textwidth}}
\vspace{20pt} 
\begin{tikzpicture}

\begin{axis}[
xmin=-7, xmax=7,
ymin=-7, ymax=7,
width=3.6cm,
height=3.6cm,
tick align=outside,
tick pos=left,
x grid style={lightgray!92.02614379084967!black},
y grid style={lightgray!92.02614379084967!black}
]


\draw[->,color=green!100,>=latex',ultra thick] (axis cs:0,0) -- (axis cs:3.5,0);
\node[color=black] (a) at (axis cs: 3,1) {\small{start}};

\draw[->,color=red!100,>=latex',ultra thick] (axis cs:0,0) -- (axis cs:-3.5,0);
\node[color=black] (a) at (axis cs: -2.68,1) {\small{goal}};

\end{axis}

\end{tikzpicture} &
\vspace{0pt} \input{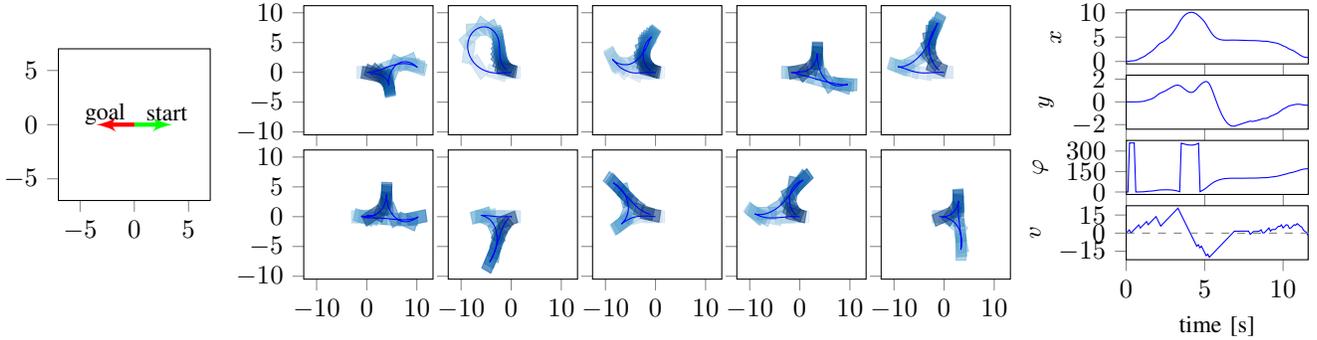} & 
\vspace{0pt} 
\begin{tikzpicture}

\begin{groupplot}[group style={group size=1 by 4,vertical sep=0.15cm}]
\nextgroupplot[
ylabel={\small{$x$}}, ylabel near ticks,
xmin=0, xmax=11.6,
ymin=-0.50481, ymax=10.60101,
width=4cm,
height=2.3cm,
tick align=outside,
tick pos=left,
xticklabels={\empty}, xlabel={\empty},
x grid style={lightgray!92.02614379084967!black},
y grid style={lightgray!92.02614379084967!black}
]
\addplot [blue, forget plot]
table {%
0 0
0.1 0
0.2 0.0375375
0.3 0.112611
0.4 0.114587
0.5 0.154101
0.6 0.231157
0.7 0.345744
0.8 0.497856
0.9 0.687443
1 0.803862
1.1 0.957611
1.2 1.14843
1.3 1.3758
1.4 1.63959
1.5 1.83056
1.6 2.05815
1.7 2.32114
1.8 2.61878
1.9 2.9494
2 3.3136
2.1 3.60601
2.2 3.82841
2.3 3.98162
2.4 4.17108
2.5 4.39715
2.6 4.66034
2.7 4.96127
2.8 5.30072
2.9 5.67958
3 6.09869
3.1 6.55876
3.2 7.06005
3.3 7.60199
3.4 8.18265
3.5 8.68668
3.6 9.11186
3.7 9.45781
3.8 9.72595
3.9 9.91916
4 10.0412
4.1 10.0962
4.2 10.0872
4.3 10.0157
4.4 9.88076
4.5 9.6777
4.6 9.3982
4.7 9.1379
4.8 8.79315
4.9 8.47403
5 8.06772
5.1 7.57456
5.2 7.00227
5.3 6.49804
5.4 5.96334
5.5 5.5249
5.6 5.1782
5.7 4.91338
5.8 4.71827
5.9 4.58024
6 4.48739
6.1 4.42908
6.2 4.39617
6.3 4.38106
6.4 4.37755
6.5 4.38071
6.6 4.3867
6.7 4.39251
6.8 4.39584
6.9 4.39488
7 4.38813
7.1 4.3806
7.2 4.37223
7.3 4.36305
7.4 4.35275
7.5 4.34142
7.6 4.32907
7.7 4.31568
7.8 4.30122
7.9 4.28519
8 4.29445
8.1 4.29535
8.2 4.28124
8.3 4.26551
8.4 4.2484
8.5 4.2293
8.6 4.20406
8.7 4.16187
8.8 4.15754
8.9 4.15386
9 4.1256
9.1 4.06965
9.2 3.9816
9.3 3.94639
9.4 3.88112
9.5 3.81486
9.6 3.74532
9.7 3.64329
9.8 3.505
9.9 3.36524
10 3.18536
10.1 3.07021
10.2 2.91533
10.3 2.7185
10.4 2.52604
10.5 2.40777
10.6 2.28124
10.7 2.11291
10.8 1.90203
10.9 1.70303
11 1.46202
11.1 1.25019
11.2 1.07915
11.3 0.92703
11.4 0.854369
11.5 0.810335
11.6 0.794763
};

\nextgroupplot[
ylabel={\small{$y$}}, ylabel near ticks,
xmin=0, xmax=11.6,
ymin=-2.4, ymax=2.35,
width=4cm,
height=2.3cm,
tick align=outside,
tick pos=left,
ytick={-2,0,2},
xticklabels={\empty}, xlabel={\empty},
x grid style={lightgray!92.02614379084967!black},
y grid style={lightgray!92.02614379084967!black}
]
\addplot [blue, forget plot]
table {%
0 0
0.1 0
0.2 -0.00146993
0.3 -0.00301689
0.4 -0.00302104
0.5 -0.00233125
0.6 0.000566791
0.7 0.00287736
0.8 0.00917049
0.9 0.0215595
1 0.0324154
1.1 0.0508629
1.2 0.079767
1.3 0.12276
1.4 0.17335
1.5 0.210497
1.6 0.253295
1.7 0.311758
1.8 0.390205
1.9 0.493911
2 0.613759
2.1 0.713995
2.2 0.790681
2.3 0.842387
2.4 0.903624
2.5 0.97321
2.6 1.04956
2.7 1.13051
2.8 1.21317
2.9 1.2937
3 1.36713
3.1 1.42714
3.2 1.46586
3.3 1.47366
3.4 1.43899
3.5 1.36473
3.6 1.26405
3.7 1.15083
3.8 1.03888
3.9 0.94123
4 0.86945
4.1 0.833171
4.2 0.839629
4.3 0.893257
4.4 0.995166
4.5 1.14234
4.6 1.32635
4.7 1.45329
4.8 1.59809
4.9 1.69126
5 1.76649
5.1 1.7949
5.2 1.74181
5.3 1.58917
5.4 1.31702
5.5 0.97846
5.6 0.605458
5.7 0.222361
5.8 -0.153167
5.9 -0.508865
6 -0.836604
6.1 -1.13128
6.2 -1.38989
6.3 -1.61085
6.4 -1.7935
6.5 -1.93768
6.6 -2.04355
6.7 -2.11138
6.8 -2.14151
6.9 -2.13424
7 -2.08986
7.1 -2.04578
7.2 -2.00186
7.3 -1.95832
7.4 -1.91388
7.5 -1.86932
7.6 -1.82479
7.7 -1.7804
7.8 -1.73624
7.9 -1.69099
8 -1.71517
8.1 -1.7174
8.2 -1.68441
8.3 -1.65021
8.4 -1.61556
8.5 -1.57954
8.6 -1.53525
8.7 -1.46671
8.8 -1.46029
8.9 -1.45514
9 -1.4177
9.1 -1.34925
9.2 -1.25231
9.3 -1.21652
9.4 -1.15216
9.5 -1.09036
9.6 -1.029
9.7 -0.94396
9.8 -0.837854
9.9 -0.741921
10 -0.631629
10.1 -0.570807
10.2 -0.496682
10.3 -0.414802
10.4 -0.348996
10.5 -0.316562
10.6 -0.286878
10.7 -0.25434
10.8 -0.224869
10.9 -0.21008
11 -0.206813
11.1 -0.219465
11.2 -0.240806
11.3 -0.267934
11.4 -0.284444
11.5 -0.295503
11.6 -0.29964
};

\nextgroupplot[
ylabel={\small{$\varphi$}}, ylabel near ticks,
xmin=0, xmax=11.6,
ymin=-17.9970054155158, ymax=377.937113725831,
width=4cm,
height=2.3cm,
tick align=outside,
tick pos=left,
ytick={0,150,300},
xticklabels={\empty}, xlabel={\empty},
x grid style={lightgray!92.02614379084967!black},
y grid style={lightgray!92.02614379084967!black}
]
\addplot [blue, forget plot]
table {%
0 0
0.1 0
0.2 359.940108310315
0.3 359.879947741826
0.4 359.879947741826
0.5 359.911460420558
0.6 0.0348612732700579
0.7 0.126569241733374
0.8 0.370366794266106
0.9 0.827230735031931
1 1.20267851902526
1.1 1.825420616975
1.2 2.75897513004936
1.3 4.06853319617832
1.4 5.36781239946263
1.5 6.15024356449328
1.6 6.89382819101407
1.7 7.97729138160645
1.8 9.46617950803341
1.9 11.4271848576482
2 13.2818492404866
2.1 14.524995768582
2.2 15.2803833256825
2.3 15.6691924854582
2.4 15.9880434984485
2.5 16.1773487539598
2.6 16.1773487539598
2.7 15.9279402257393
2.8 15.3685042345736
2.9 14.4371040428089
3 13.07082888454
3.1 11.2049918246964
3.2 8.7728432801456
3.3 5.70551945349069
3.4 1.93115424848844
3.5 358.198889570912
3.6 354.627070676067
3.7 351.340011805401
3.8 348.463190716049
3.9 346.128387700891
4 344.474831504144
4.1 343.64576157459
4.2 343.795303559119
4.3 345.087323387139
4.4 347.696000228369
4.5 351.809837197409
4.6 357.633953184913
4.7 2.38485660814073
4.8 8.89293523400502
4.9 14.6791787112517
5 21.8711168430723
5.1 30.4688642210259
5.2 40.472363549333
5.3 49.1144578606298
5.4 58.3683565055672
5.5 66.3788794392913
5.6 73.262394389993
5.7 79.1260444653618
5.8 84.0683784061603
5.9 88.1827883329947
6 91.5552179151348
6.1 94.2681730750792
6.2 96.3990031151708
6.3 98.0216195909812
6.4 99.2070692691069
6.5 100.022961169373
6.6 100.535758396015
6.7 100.809632222088
6.8 100.90703504726
6.9 100.889273355611
7 100.81765363122
7.1 100.782130247922
7.2 100.782130247922
7.3 100.81765363122
7.4 100.890419271201
7.5 101.001000125662
7.6 101.1493961946
7.7 101.336180435813
7.8 101.560779891504
7.9 101.832361886396
8 101.664485252423
8.1 101.646723560774
8.2 101.940077951881
8.3 102.279841924394
8.4 102.662577731541
8.5 103.102036360406
8.6 103.697912467342
8.7 104.710901849134
8.8 104.815753125643
8.9 104.906853415068
9 105.630499110319
9.1 107.080655289795
9.2 109.36045435662
9.3 110.234214994195
9.4 111.829329495839
9.5 113.406682305834
9.6 115.020704414718
9.7 117.332589118071
9.8 120.366400643288
9.9 123.316560330417
10 126.989219797205
10.1 129.255840834743
10.2 132.244388694145
10.3 135.954863375413
10.4 139.494596633731
10.5 141.629437378388
10.6 143.891474753565
10.7 146.875438950606
10.8 150.581329969512
10.9 154.054600123595
11 158.249797099543
11.1 161.943083046956
11.2 164.943090062261
11.3 167.632553952605
11.4 168.929730400782
11.5 169.719839200267
11.6 170.000015562086
};

\nextgroupplot[
xlabel={\small{time [s]}}, xlabel near ticks,
ylabel={\small{$v$}}, ylabel near ticks,
xmin=0, xmax=11.6,
ymin=-22.0409244, ymax=22.7101644,
width=4cm,
height=2.3cm,
tick align=outside,
tick pos=left,
ytick={-15,0,15},
x grid style={lightgray!92.02614379084967!black},
y grid style={lightgray!92.02614379084967!black}
]
\addplot [blue, forget plot]
table {%
0 0
0.1 1.35135
0.2 2.7027036
0.3 0.07112376
0.4 1.4224752
0.5 2.7738252
0.6 4.125168
0.7 5.476536
0.8 6.827868
0.9 4.196304
1 5.547636
1.1 6.899004
1.2 8.250372
1.3 9.601704
1.4 6.97014
1.5 8.310996
1.6 9.651852
1.7 11.003184
1.8 12.354552
1.9 13.705884
2 11.07432
2.1 8.44272
2.2 5.811156
2.3 7.162524
2.4 8.513856
2.5 9.865224
2.6 11.216556
2.7 12.567924
2.8 13.919256
2.9 15.270624
3 16.621956
3.1 17.973324
3.2 19.324656
3.3 20.676024
3.4 18.044424
3.5 15.41286
3.6 12.781296
3.7 10.149696
3.8 7.518132
3.9 4.886532
4 2.2549644
4.1 -0.376614
4.2 -3.0081924
4.3 -5.63976
4.4 -8.27136
4.5 -10.902924
4.6 -9.551592
4.7 -12.183156
4.8 -10.831824
4.9 -13.463388
5 -16.094952
5.1 -18.726552
5.2 -17.375184
5.3 -20.006784
5.4 -18.655416
5.5 -17.304084
5.6 -15.952716
5.7 -14.601384
5.8 -13.250016
5.9 -11.898684
6 -10.547316
6.1 -9.195948
6.2 -7.844616
6.3 -6.493248
6.4 -5.141916
6.5 -3.790548
6.6 -2.4392088
6.7 -1.0878588
6.8 0.26349408
6.9 1.6148448
7 1.609578
7.1 1.6098156
7.2 1.6015572
7.3 1.640988
7.4 1.6522956
7.5 1.658196
7.6 1.6611624
7.7 1.6613172
7.8 1.7113572
7.9 -0.9202212
8 -0.08534628
8.1 1.2660048
8.2 1.3217904
8.3 1.3498164
8.4 1.4156856
8.5 1.7590644
8.6 2.756808
8.7 0.26318376
8.8 0.2130336
8.9 1.5643836
9 2.9157372
9.1 4.26708
9.2 1.6355088
9.3 2.9868588
9.4 2.9520108
9.5 3.0217032
9.6 4.327812
9.7 5.679144
9.8 5.523372
9.9 6.87474
10 4.24314
10.1 5.594508
10.2 6.94584
10.3 6.627132
10.4 3.995568
10.5 4.234788
10.6 5.586156
10.7 6.937488
10.8 6.501924
10.9 7.853292
11 6.91398
11.1 5.61618
11.2 5.034564
11.3 2.4278184
11.4 1.4792832
11.5 0.5250096
11.6 -2.1065688
};
\addplot [black!50, dashed, forget plot]
table {%
0 0
11.7 0
};

\end{groupplot}

\end{tikzpicture}
\end{tabular}
\vspace{-0.2cm}
\caption{Experiment 1. (Left) mission formulation: a motion from ``start''- to ``goal''-pose (both at 0-velocity) is sought. (Center) Planar trajectories for 10 different initial random seeds. In all cases a NN-[5,2,2] is employed. More quantifications are: $\bar{\tau}_{a}=0.034$s, $v^\text{min}/\bar{v}/v^\text{max}=-29/-1.1/26$km/h and $P^\text{min}/\bar{P}/P^\text{max}=21.6/26.7/33.4$m. In 9 out of 10 cases 3-point-steering naturally evolves. (Right) Vehicle states for the grid view at the top-left. }
\label{fig:exrrt_planarGrid}
\vspace{-0.4cm}
\end{figure*}

\section{Solution Desription\label{sec_problem_soln}}

\subsection{Closed-loop algorithm\label{subsec_closedLoopAlg}}

This paper proposes the closed-loop control architecture visualized in Fig. \ref{fig:closed_loop} and Algorithm \ref{alg_clLoop} for closed-loop control of transportation missions in the autonomous vehicles context. The main submodules are discussed below.

\vspace{-0.1cm}
\begin{algorithm}
\begin{small}
\DontPrintSemicolon
\textbf{Mission Start:} $\mathcal{M}$, $z_0$ and $t=0$; Sect. \ref{subsec_step1}.\;
\While{mission not yet completed}
{
Obtain measurements: $m_t^z$, $m_t^\text{ext}$, $m_t^\text{ref}$; Sect. \ref{subsec_obtainMeasurements}.\;
Determine goal setpoint: $\zeta_t^\text{goal}$; Sect. \ref{subsec_targetPt}.\;
Determine control action: $a_t$; Sect. \ref{subsec_at}.\;
Apply control action $a_t$ to the vehicle.\;
Determine if mission is yet completed.\;
Wait until the next sampling time: $t=t+1$.\;
} 
%
\caption{Closed-loop control algorithm}\label{alg_clLoop}
\end{small}
\end{algorithm}
\vspace{-0.5cm}

\subsection{Step 1: Mission start\label{subsec_step1}}

A priori \emph{mission data} is summarized by $\mathcal{M}$, output of a high-level route planner (not subject of this paper). It comprises at least 1 goal setpoint, i.e., the final goal pose described by at least planar location, heading and velocity. It may also describe a reference trajectory, i.e., a sequence of setpoints uniformly or non-uniformly spatially distributed. Ultimately, it may additionally include obstacle information, e.g., map-based information about road-bounds.  In general, all data may also be available as sets accounting for uncertainty rather than as point-data only. A transportation mission is completed (Step 7 of Alg. \ref{alg_clLoop}) once the final goal setpoint is reached (e.g., a parking position) within a specified tolerance.  

The initial vehicle state is denoted by $z_0$. It may not necessarily be fully observable. Thus, full state vector information may not be available to the control algorithm.

\subsection{Step 3: Obtaining measurements\label{subsec_obtainMeasurements}}

According to Fig. \ref{fig:closed_loop}, it is distinguished between proprio- ($m_t^z$), exteroceptive ($m_t^\text{ext}$), and reference measurements obtained from a high-level route-planner ($m_t^\text{ref}$) . 

Measurements $m_t^z$ can represent any subset of $z_t$ arbitrarily perturbed by noise. In the simplest case, $m_t^z=z_t$.

Measurements $m_t^\text{ext}$ are assumed to be available (e.g., from lidar and camera sensors) as a set of \emph{obstacle point data} in the EV-aligned coordinate system spanned by $\xi$- and $\eta$-coordinates.  Every of the $N_\text{obstPts}$ obstacle points is assumed as a 4D vector (planar position, heading and velocity). Since vehicles are throughout assumed as ground-contacted systems, projection of all relevant 3D obstacle data to the 2D plane permits w.l.o.g. 3D-collision-free trajectory planning. Projections may also account for vehicle height, roll-angle and 3D obstacles.

Reference route data $m_t^\text{ref}$  may become available as a subset (for finite horizon planning) of preliminary mission data $\mathcal{M}$ indexed spatially along its path. Thus, the first setpoint of $m_t^\text{ref}$ shall coincide with the reference spatially closest to the absolute $(x,y)$-coordinates of the EV\footnote{In implementation practice, a discrete space-index is used to track the EV's progress and position along the reference path in case this is available.}.  Alternatively, variable $m_t^\text{ref}$ is also employed to account for all potentially updated mission data.

\subsection{Step 4: Goal setpoint selection\label{subsec_targetPt}}

According to Step 4 of Alg. \ref{alg_clLoop} (intermediate) goal setpoints are set at every sampling time $t$ and described by location, heading and velocity information, i.e., 
\begin{equation}
\zeta_t^\text{goal} = \begin{bmatrix} x_t^\text{goal} & y_t^\text{goal} & \varphi_t^\text{goal} & v_t^\text{goal} \end{bmatrix}.
\end{equation}
Two comments are made. First, the critical importance of  $\zeta_t^\text{goal} $ for Alg. \ref{alg_clLoop} is emphasized. The fundamental nature of the proposed method is sampling-based. Therefore, $\zeta_t^\text{goal}$ is selected ideally such that it is reached over the prediction horizon in Step 5 of Alg. \ref{alg_clLoop}. However, here no guarantee can be given that it actually will be reached. Efforts undertaken to address this issue heuristically are discussed further in Sect. \ref{sec_expts}, and in Sect. \ref{sec_discussion} discussing limitations of the presented method and outlining ongoing work in this direction.

Second, conceptually the selection of $\zeta_t^\text{goal}$ is to be interpreted as a strategic upstream decision layer \emph{preceding} the control signal generator, which is the main subject of this paper and which is responsible for motion planning of the current vehicle state to $\zeta_t^\text{goal}$. Considerations for $\zeta_t^\text{goal}$-design are (i) traffic rules (e.g., setting a traffic light stopping position, a headway position subject to road speed limits, or a leading vehicle for adaptive cruise control as setpoints), and (ii) a recursive logic based on success of reaching $\zeta_t^\text{goal}$ over the prediction horizon at the last sampling time.

\subsection{Step 5:  Determining control action \label{subsec_at}}

In \cite{plessen2019automating}, the TSHC-algorithm is proposed for \emph{offline} encoding of multiple motion primitives in a neural network. For information, a single motion primitive may, e.g., be a left-turn connecting a specific start and goal vehicle pose (described by at least planar position, heading and velocity).

In this paper, the same TSHC-algorithm is also applied, however, for \emph{online} motion planning. Therefore, at every sampling time $t$ only \emph{one} trajectory from the current vehicle state to the designated goal setpoint $\zeta_t^\text{goal}$ is sought online (instead of multiple motion primitives encoded offline as in \cite{plessen2019automating,plessen2018encoding}). By construction, TSHC samples in the parameter space of a neural network that is parameterizing controller C in Fig. \ref{fig:closed_loop}. The method is suitable for GPU-acceleration. Several comments are made.

First, under the assumption of employing a fixed vehicle model for the EV (e.g., a kinematic or dynamic one) for the forward simulation of the EV-motion over $t,\dots,t+H$, such vehicle model and all its (constant) hyperparameters can be defined \text{directly on} the GPU (without the need for passing between host and GPU), since this data is not changing.  In contrast, obstacle information at time $t$ must be passed to the GPU, since these measurement data are changing every $t$. In this paper, the space environment that is prohibited from being trespassed by the EV for collision avoidance is modeled as a set of obstacle points. This data stems from both static and dynamic obstacles. For dynamic motion planning the environment must then be extrapolated over a prediction horizon $H$, i.e., $\{\hat{m}_{t+h}^\text{ext}\}_{h=0}^H$ with $\hat{m}_{t+0}^\text{ext}=m_{t}^\text{ext}$. In more detail,  in this paper the following is therefore done.
\begin{enumerate}
\item[(i)] The obstacle points data vector $m_t^\text{ext}$ is assumed to contain planar location, heading and velocity information. In practice (and not subject of this paper), only the 2 last relative location positions may be retrieved, based on which heading and velocity must then be estimated, for example, using sensor fusion and model-based estimation techniques such as Kalman filtering.
\item[(ii)] The movement of all obstacle points is extrapolated under the assumptions of \emph{constant velocity} over the prediction horizon $H$. This is done for a practical reason. Obstacle point locations are extrapolated linearly. In contrast to alternative history-dependent nonlinear extrapolation techniques (e.g., constant acceleration with velocity capping), this enables fast collision checks when gridding over all obstacle points at every prediction time $t+h,~\forall h=0,\dots,H$.
\item[(iii)] Only $m_t^\text{ext}$ is passed to the GPU at every $t$. Extrapolating $\{\hat{m}_{t+h}^\text{ext}\}_{h=0}^H$ over the prediction horizon is then implemented directly on the GPU.
\item[(iv)] Extrapolating the movement of obstacle points is done only \emph{once} at time $t$ under the constant velocity assumption, but \emph{not} updated again throughout the simulation horizon $\{t+1,\dots,t+H\}$. In general, when the EV is moving throughout the duration of the simulation horizon, new obstacle points (especially from road bounds) from offline stored maps may become available. However, all of the latter are dismissed. The dismissal is done for computational efficiency. Motion planning can be GPU-accelerated based thereon and without requiring any additional calls to offline stored maps. To further mitigate the relevance of any new obstacle points, the goal setpoint $\zeta_t^\text{goal}$ is always selected \emph{within} or \emph{on} the boundary of the (e.g., rectangular) range field visible at time $t$. 
\end{enumerate}
\vspace{-0.2cm}

\setlength\figureheight{6.5cm}
\setlength\figurewidth{6.5cm}
\vspace{0.3cm}
\begin{figure}
\centering%
\begin{tikzpicture}

\definecolor{color1}{rgb}{0.280392156862745,0.338158274815817,0.985162233467507}
\definecolor{color0}{rgb}{0,0,0}
\definecolor{color3}{rgb}{0.166666666666667,0.866025403784439,0.866025403784439}
\definecolor{color2}{rgb}{0.0607843137254902,0.636474236147141,0.941089252501372}
\definecolor{color5}{rgb}{0.613725490196078,0.984086337302604,0.641213314833578}
\definecolor{color4}{rgb}{0.386274509803922,0.984086337302604,0.767362681448697}
\definecolor{color7}{rgb}{1,0.636474236147141,0.338158274815817}
\definecolor{color6}{rgb}{0.833333333333333,0.866025403784439,0.5}
\definecolor{color9}{rgb}{1,0,0}
\definecolor{color8}{rgb}{1,0.338158274815818,0.17162567916636}

\begin{axis}[
xlabel={\small{$x$ [m]}}, xlabel near ticks,
ylabel={\small{$y$ [m]}}, ylabel near ticks,
xmin=-2.45043, xmax=51.45903,
ymin=-1, ymax=1,
width=8.5cm,
height=2.5cm,
tick align=outside,
tick pos=left,
x grid style={lightgray!92.02614379084967!black},
y grid style={lightgray!92.02614379084967!black}
]
\addplot [color0, line width=1pt, forget plot]
table {%
0 0
1.38889 0.0271606
2.81505 0.0548322
4.27929 0.0546085
5.78079 0.0530344
7.3198 0.0213239
8.89598 -0.0142187
10.5104 -0.0190506
12.162 -0.0190503
13.8512 -0.0520832
15.4669 -0.0902116
17.1191 -0.161548
18.6978 -0.235312
20.3154 -0.279223
21.9704 -0.319103
23.5531 -0.326286
25.1729 -0.329509
26.8303 -0.300394
28.5248 -0.264531
30.2576 -0.261745
32.0276 -0.267123
33.7246 -0.239094
35.348 -0.205543
36.899 -0.203819
38.3766 -0.205291
39.7811 -0.179224
41.1122 -0.154433
42.4812 -0.155707
43.7768 -0.15635
45.11 -0.130941
46.3698 -0.108104
47.6662 -0.059234
48.9943 0.041071
};
\addplot [color1, line width=1pt, forget plot]
table {%
0 0
1.38889 0.0271606
2.81505 0.0548322
4.27929 0.0546085
5.78079 0.0530344
7.3198 0.0213239
8.89458 -0.0759399
10.503 -0.214926
12.0379 -0.352188
13.5025 -0.454346
15.0085 -0.498855
16.4426 -0.452643
17.9097 -0.345676
19.3006 -0.214117
20.6184 -0.0895222
21.9761 0.0121042
23.375 0.0626456
24.8134 0.030041
26.2854 -0.0634035
27.6814 -0.182394
29.0042 -0.295171
30.2563 -0.377294
31.4382 -0.41085
32.5466 -0.427395
33.5817 -0.447059
34.544 -0.446523
35.5437 -0.412937
36.5801 -0.369432
37.6538 -0.329808
38.7657 -0.310533
39.9158 -0.330986
41.1015 -0.366978
42.3243 -0.400441
43.5851 -0.410282
44.8835 -0.372111
46.2157 -0.258137
47.5789 -0.0912668
48.9749 0.106338
};
\addplot [color2, line width=1pt, forget plot]
table {%
0 0
1.38889 -0.0271606
2.81505 -0.0548321
4.27929 -0.0546085
5.78079 -0.053034
7.31986 -0.0815171
8.89611 -0.113754
10.5105 -0.1152
12.1622 -0.111737
13.8514 -0.141227
15.5778 -0.178346
17.3424 -0.181778
19.1442 -0.17611
20.873 -0.204477
22.5282 -0.239245
24.2217 -0.241697
25.9523 -0.237112
27.61 -0.265137
29.1941 -0.297624
30.8143 -0.299166
32.3611 -0.295883
33.8347 -0.263938
35.2349 -0.232138
36.5626 -0.227951
37.8172 -0.271847
38.9972 -0.331458
40.2147 -0.389236
41.4708 -0.424251
42.7656 -0.412254
43.9862 -0.380668
45.2442 -0.351264
46.5391 -0.295648
47.7611 -0.244943
49.005 -0.217675
};
\addplot [color3, line width=1pt, forget plot]
table {%
0 0
1.38889 0.0271606
2.81505 0.0548322
4.27929 0.0546085
5.78079 0.0530344
7.31986 0.0815178
8.7855 0.111493
10.1786 0.112741
11.609 0.114162
12.9662 0.142053
14.2502 0.167674
15.5721 0.168202
16.9315 0.117001
18.3263 0.0386917
19.7584 -0.0416455
21.2296 -0.0953471
22.6294 -0.0902258
24.0656 -0.0569043
25.539 -0.0219676
26.9401 -0.016148
28.3787 -0.0665178
29.8526 -0.148618
31.3634 -0.234388
32.9133 -0.292
34.5025 -0.285498
36.1275 -0.238486
37.679 -0.188709
39.1583 -0.170187
40.5648 -0.209197
41.8966 -0.272459
43.2654 -0.336205
44.6728 -0.374187
46.1186 -0.356452
47.6003 -0.3074
49.0086 -0.25855
};
\addplot [color4, line width=1pt, forget plot]
table {%
0 0
1.38889 0.0271606
2.81505 0.0548322
4.27929 0.0546085
5.78079 0.0530344
7.31986 0.0815178
8.89611 0.113755
10.5105 0.115202
12.1622 0.111738
13.8513 0.141227
15.5772 0.178332
17.2307 0.181545
18.8109 0.179021
20.4285 0.208069
22.0828 0.242809
23.6646 0.245093
25.1731 0.243443
26.6084 0.213803
28.0809 0.182664
29.5914 0.180263
31.1388 0.240783
32.7207 0.340235
34.229 0.438977
35.6663 0.504873
37.0327 0.513259
38.3255 0.446356
39.5411 0.341089
40.7913 0.214511
42.0786 0.0867913
43.4055 -0.0187135
44.7734 -0.0750575
46.1808 -0.0514813
47.6217 0.029544
48.9869 0.134767
};
\addplot [color5, line width=1pt, forget plot]
table {%
0 0
1.38889 -0.0271606
2.81396 -0.110679
4.27366 -0.226018
5.77039 -0.345633
7.19648 -0.431579
8.55253 -0.459595
9.83608 -0.412206
11.043 -0.326175
12.2854 -0.219831
13.5651 -0.113044
14.8844 -0.0288924
16.2408 0.059112
17.6364 0.122289
19.0693 0.187212
20.541 0.22508
22.0499 0.262389
23.5971 0.270382
25.1816 0.275313
26.8037 0.24864
28.3523 0.218369
29.939 0.218382
31.563 0.222629
33.114 0.196356
34.5915 0.168209
36.107 0.168975
37.6596 0.172261
39.1393 0.146459
40.5455 0.12037
41.9897 0.121819
43.4706 0.124216
44.8785 0.0989638
46.2131 0.0748778
47.5851 0.0769454
48.9938 0.133681
};
\addplot [color6, line width=1pt, forget plot]
table {%
0 0
1.38889 -0.0271606
2.81505 -0.0548321
4.16863 -0.0546254
5.44889 -0.00512907
6.76492 0.0671481
8.11827 0.139919
9.51056 0.187506
10.942 0.180527
12.4097 0.143388
13.8041 0.106779
15.126 0.0979263
16.4852 0.0873411
17.8818 0.0491493
19.3156 0.00999944
20.7861 -0.0589476
22.2938 -0.131157
23.8404 -0.174944
25.3137 -0.213627
26.8254 -0.223749
28.3743 -0.231707
29.9609 -0.208834
31.4759 -0.183042
33.0291 -0.186973
34.6195 -0.194397
36.1392 -0.171775
37.6932 -0.146042
39.1745 -0.150478
40.6904 -0.156738
42.1332 -0.134482
43.5027 -0.112537
44.9072 -0.117495
46.2384 -0.122275
47.6034 -0.100486
49.0011 -0.0788554
};
\addplot [color7, line width=1pt, forget plot]
table {%
0 0
1.38889 0.0271606
2.81505 0.0548322
4.27929 0.0546085
5.78078 0.0530344
7.3198 0.081517
8.89572 0.113747
10.5098 0.115192
12.1596 0.111739
13.847 0.141204
15.5715 0.178246
17.2236 0.181423
18.8023 0.178903
20.4185 0.207928
22.0718 0.242615
23.6526 0.244869
25.16 0.243213
26.7049 0.271727
28.2868 0.30413
29.7963 0.305529
31.2324 0.304667
32.706 0.332598
34.2167 0.36279
35.6549 0.363407
37.0197 0.363267
38.422 0.335701
39.7508 0.309532
41.1167 0.255903
42.4093 0.20575
43.7405 0.180154
45.1088 0.152527
46.4046 0.151706
47.7377 0.14892
48.9977 0.170927
};
\addplot [color8, line width=1pt, forget plot]
table {%
0 0
1.38889 -0.0271606
2.81505 -0.0548321
4.27929 -0.0546085
5.67016 -0.05315
7.09859 -0.0795856
8.56426 -0.107295
9.95737 -0.10639
11.2771 -0.105662
12.6344 -0.131456
14.029 -0.157129
15.351 -0.155614
16.5996 -0.155546
17.8857 -0.180627
19.2092 -0.204333
20.46 -0.202278
21.6375 -0.156689
22.7407 -0.0992246
23.8813 -0.0445246
25.0605 -0.0110653
26.2784 -0.0204367
27.5329 -0.0482265
28.7141 -0.071995
29.8226 -0.0726228
30.8581 -0.0368852
31.8195 0.00532046
32.8184 0.0430578
33.8557 0.0619432
34.9311 0.0448754
36.0433 0.0155702
37.1927 -0.0100989
38.2695 -0.0130871
39.2733 0.0188933
40.2032 0.0564169
41.1706 0.0891047
42.1762 0.103408
43.2198 0.0831505
44.3001 0.0518157
45.4168 0.024409
46.5705 0.0186586
47.7608 0.0552278
48.985 0.157868
};
\addplot [color9, line width=1pt, forget plot]
table {%
0 0
1.38889 -0.0271606
2.81505 -0.0548321
4.27929 -0.0546085
5.67016 -0.05315
7.09859 -0.0795856
8.56426 -0.107295
9.95737 -0.10639
11.2771 -0.105662
12.6344 -0.078371
13.9183 -0.0533176
15.2403 -0.0533738
16.5995 -0.0519487
17.8857 -0.0757518
19.2092 -0.0981431
20.46 -0.0948451
21.6375 -0.0941955
22.8526 -0.117286
24.105 -0.137852
25.2847 -0.134156
26.391 -0.134096
27.5349 -0.156403
28.7162 -0.175215
29.8248 -0.171191
30.8601 -0.131109
31.8213 -0.0848691
32.8201 -0.0429403
33.8572 -0.0197026
34.9316 0.00981319
35.9334 0.0177408
36.9725 0.0317433
38.0494 0.0251972
39.1637 0.023453
40.2049 0.0421841
41.1729 0.0547475
42.1786 0.0481351
43.2216 0.0470227
44.3021 0.0669992
45.4197 0.0841414
46.5753 0.0804791
47.7681 0.0820628
48.9977 0.109117
};
%
%
%

\end{axis}

\end{tikzpicture}
\vspace{-0.5cm}\caption{Experiment 2 for NN-[5,2,2]. Trajectories for 10 different initial random seeds are displayed. Because of the sampling-based nature of the algorithm lateral deviations result. See also Table \ref{tab_ex2}.}
\label{fig_Ex1_522_planarOverlay}
\vspace{-0.5cm}
\end{figure}
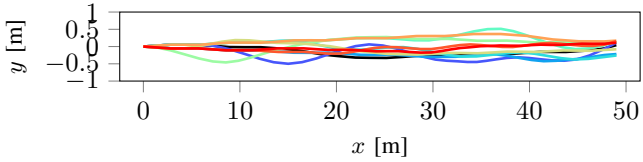
  

Second, the employed method for collision-checks (implemented on the GPU) is briefly discussed.  In general, the EV can be modeled as an arbitrarily refined polytope (convex or non-convex). Then, a collision with obstacle points boils down to a multidimensional linear inequality check. For example, for typical 2D navigation in the plane it is
\begin{equation}
A(\varphi_{t+h})\left( \begin{bmatrix} \hat{m}_{t+h}^\text{ext,x} \\ \hat{m}_{t+h}^\text{ext,y} \end{bmatrix}_{j} - \begin{bmatrix} x_{t+h} \\ y_{t+h} \end{bmatrix} \right) < b(\varphi_{t+h}),
\label{eq_Apsietc_collAvoidIneqs}
\end{equation}
$\forall h=0,\dots,H$, $\forall j=1,\dots,N_\text{obstPts}$, and where $A(\varphi_{t+h})$ and $b(\varphi_{t+h})$ are algebraic functions of the vehicle dimensions defining the polygon for the collision check. 


Third, due to the sampling-nature of proposed algorithm and typical small sampling times of the closed-cloop control system, in general, no guarantee about generating a trajectory actually reaching the goal setpoint can here be given (see Sect. \ref{sec_concl} for ongoing work in this direction). For example, (i) the goal setpoint may be selected as too difficult to be reachable within $t,\dots,t+H$, or (ii) the number of samples is too small (due to computational constraints), or (iii) algorithm hyperparameters may be set unsuitably to generate such trajectory. However, even if not reaching the final goal setpoint, TSHC is designed to at least return a collision-free trajectory. In the extreme case, again due to the sampling nature of the algorithm, no guarantee abound finding any collision-free trajectory may be given (e.g., due to a cluttered environment too difficult to navigate in). The larger the number of samples that can be generated by the GPU, the larger the likelihood of generating collision-free trajectories reaching the designated goal setpoint.

Fourth, for clarity it is summarized what (i) must be passed from host to GPU as part of Step 5 of Alg. \ref{alg_clLoop}  at every $t$, and (ii) what hyperparameters and models can be defined directly on the GPU since being invariant over $t$. For the former, these are $\theta,~m_t^z,~a_{t-1},~\zeta_t^\text{goal}$, $m_t^\text{ext}$ and the current random seed. For the latter, these are the vehicle system model, the controller parametrization structure (here multilayer perceptrons), the method to update $\sigma_\text{pert}$ (here randomly as in \cite{plessen2018encoding}), and all constant hyperparameters from Table \ref{tab_hyperparameters}.

Fifth, $\theta$ is initialized at every $t$ for $i_\text{restart}=1,\dots,N_\text{restarts}$ as follows:
\begin{equation*}
\theta(i_\text{restart})=\begin{cases} \theta_{t-1}^\star,~&\text{if}~i_\text{restart}=1,\\
\theta(i_\text{restart}-1),~&\text{if}~2\leq i_\text{restart}\leq N_\text{restarts},
\end{cases}
\end{equation*} 
where $\theta_{t-1}^\star$ is the best parametrization that last sampled a trajectory reaching the goal setpoint in prediction.

Sixth, in this paper the feature vector is selected as 
\begin{small}
\begin{align}
s_{t+h} = &~\large[ \frac{\xi_{t}^\text{goal}-\xi_{t+h}}{\Delta \xi},~ \frac{\eta_{t}^\text{goal}-\eta_{t+h}}{\Delta \eta},\dots \label{eq_def_stph}\\
 & \hspace{0.7cm}\dots,~\frac{\varphi_{t}^\text{goal}-\varphi_{t+h}}{\Delta \varphi},~\frac{v_{t}^\text{goal}-v_{t+h}}{\Delta v},~a_{t-1+h}[0] \Large],\notag
\end{align}
\end{small}
$\forall h=0,\dots,H$, and where the first 4 elements denote deviations of the EV w.r.t. $\zeta_t^\text{goal}$ over the prediction horizon in the EV-aligned coordinate system, and where $a_{t-1+h}[0]$ represents the NN-output related to steering control at the subscripted time-instances. All normalization constants are indicated by $\Delta(\cdot)$ to normalize elements for $s_t$ approximately to the range of $\pm 1$. See Table \ref{tab_hyperparameters} for numerical values.

Ultimately, the final control is obtained as the mapping
\begin{equation}
a_t = \mathcal{X}(s_t,\theta^\star),\label{eq_def_at_mathcalXetc}
\end{equation}
where $\theta^\star$ are the optimal parameters returned by TSHC.

\setlength\figureheight{6.5cm}
\setlength\figurewidth{6.5cm}
\begin{figure*}
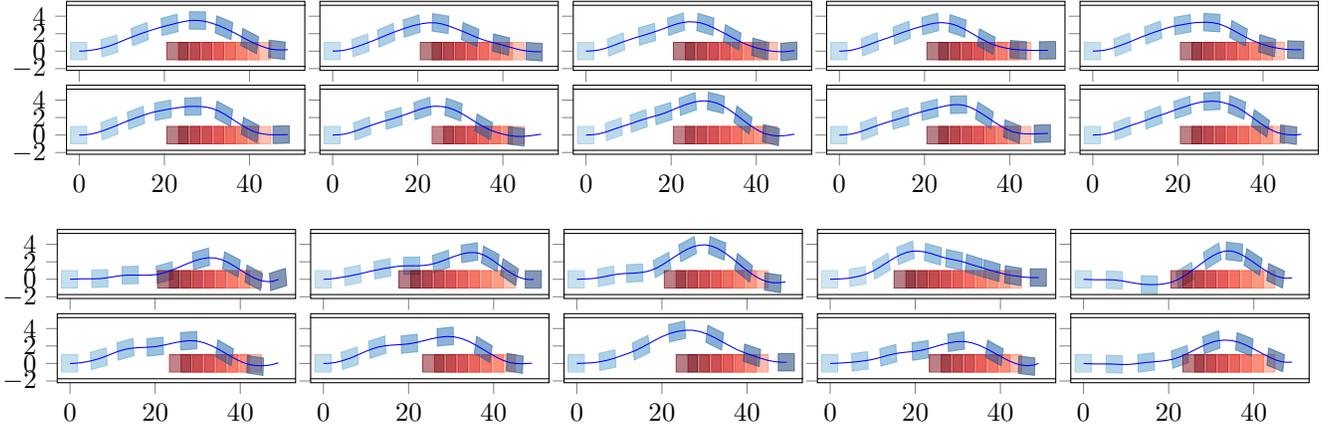

\vspace{0.2cm}
\centering%
\input{Ex6_planarGrid.tex}\\[10pt]
\input{Ex4_planarGrid.tex}
\vspace{-0.cm}\caption{Experiment 3. Trajectories for 10 different initial random seeds are displayed. The EV is driving from left to right, while the obstacle vehicle (red) is driving from right to left. (TOP 10) The case of implicitly considering the dynamic obstacle by setting an auxiliary setpoint in the neighboring lane to ensure obstacle avoidance. Obstacle points are only perceived from the static road bounds. (BOTTOM 10) The case of explicitly considering obstacle points of the dynamic obstacle. There are 4 obstacle points perceived from the obstacle vehicle and 16 from the road bounds. See Sect. \ref{subsec_Ex4and6}.}
\label{fig:ex6_planarGrid}
\vspace{-0.2cm}
\end{figure*}

\setlength\figureheight{6.5cm}
\setlength\figurewidth{6.5cm}
\begin{figure}
\centering%
\begin{tikzpicture}

\definecolor{color1}{rgb}{0.280392156862745,0.338158274815817,0.985162233467507}
\definecolor{color0}{rgb}{0,0,0}
\definecolor{color3}{rgb}{0.166666666666667,0.866025403784439,0.866025403784439}
\definecolor{color2}{rgb}{0.0607843137254902,0.636474236147141,0.941089252501372}
\definecolor{color5}{rgb}{0.613725490196078,0.984086337302604,0.641213314833578}
\definecolor{color4}{rgb}{0.386274509803922,0.984086337302604,0.767362681448697}
\definecolor{color7}{rgb}{1,0.636474236147141,0.338158274815817}
\definecolor{color6}{rgb}{0.833333333333333,0.866025403784439,0.5}
\definecolor{color9}{rgb}{1,0,0}
\definecolor{color8}{rgb}{1,0.338158274815818,0.17162567916636}

\begin{axis}[
xlabel={\small{$x$ [m]}}, xlabel near ticks,
ylabel={\small{$y$ [m]}}, ylabel near ticks,
xmin=-3, xmax=53,
ymin=-2.1, ymax=5.6,
width=4.25cm,
height=3cm,
tick align=outside,
tick pos=left,
x grid style={lightgray!92.02614379084967!black},
y grid style={lightgray!92.02614379084967!black}
]
\addplot [black, mark=*, mark size=3, mark options={solid}, line width=1pt, forget plot]
table {%
-10 -1.75
60 -1.75
};
\addplot [black, mark=*, mark size=3, mark options={solid}, line width=1pt, forget plot]
table {%
-10 5.25
60 5.25
};
\addplot [color0, line width=1pt, forget plot]
table {%
0 0
1.38889 0.0271606
2.81505 0.0548322
4.27929 0.0546085
5.67016 0.0531504
7.09859 0.0795862
8.45264 0.158268
9.84169 0.26455
11.2679 0.373537
12.7337 0.456761
14.2401 0.481807
15.7834 0.472705
17.3639 0.460222
18.9821 0.479085
20.6366 0.568252
22.3205 0.770961
24.0272 1.06626
25.7604 1.41789
27.5283 1.78602
29.2314 2.10607
30.8755 2.34105
32.4602 2.46154
33.9781 2.44551
35.4157 2.2787
36.7605 2.00999
38.0151 1.68386
39.1863 1.33554
40.2819 0.992272
41.3086 0.674536
42.2713 0.397124
43.2784 0.143189
44.3332 -0.0687688
45.437 -0.2169
46.5875 -0.275752
47.7769 -0.216469
48.9893 -0.00760438
};
\addplot [color1, line width=1pt, forget plot]
table {%
0 0
1.38889 0.0271606
2.81396 0.110679
4.16334 0.217301
5.54996 0.325959
6.97181 0.465398
8.4307 0.609056
9.83731 0.775394
11.171 0.93342
12.5452 1.06905
13.9564 1.20872
15.4078 1.32377
16.7861 1.43193
18.204 1.51535
19.5518 1.54143
20.8259 1.54302
22.1374 1.54697
23.4862 1.57742
24.8712 1.662
26.2876 1.83195
27.7317 2.06448
29.2066 2.3343
30.609 2.59286
31.9443 2.81216
33.2152 2.97094
34.4212 3.05409
35.5583 3.05222
36.6201 2.96148
37.5984 2.78335
38.49 2.56138
39.2989 2.32121
40.1323 2.0486
40.995 1.7499
41.8909 1.43376
42.8241 1.11105
43.7993 0.795012
44.8213 0.501688
45.8942 0.250354
47.0199 0.0639203
48.1968 -0.0307885
49.4167 -0.012886
};
\addplot [color2, line width=1pt, forget plot]
table {%
0 0
1.38889 0.0271606
2.81505 0.0548322
4.16761 0.107546
5.55537 0.215234
6.97736 0.353258
8.43638 0.495464
9.8253 0.603471
11.1451 0.654393
12.3917 0.680871
13.6755 0.710889
14.9962 0.767623
16.3521 0.877657
17.6281 1.05572
18.926 1.33572
20.1344 1.6664
21.2601 2.01471
22.4162 2.38938
23.6087 2.77248
24.8447 3.14303
26.1309 3.47647
27.472 3.74397
28.869 3.91205
30.3159 3.94244
31.6857 3.80411
33.07 3.52808
34.3552 3.16407
35.5474 2.75422
36.7592 2.292
37.9973 1.79599
39.2705 1.28729
40.5659 0.798872
41.8931 0.356616
43.2534 -0.0094583
44.6772 -0.274462
46.1605 -0.397831
47.5731 -0.395546
49.0174 -0.307287
};
\addplot [color3, line width=1pt, forget plot]
table {%
0 0
1.38889 0.0271606
2.70443 0.0526859
3.94642 0.10109
5.22394 0.198238
6.53636 0.319483
7.77146 0.481098
8.92511 0.696491
9.9993 0.930148
11.0977 1.22771
12.1162 1.5346
13.1667 1.86282
14.2544 2.19762
15.3846 2.5215
16.5629 2.81369
17.7932 3.04966
19.0765 3.2007
20.4082 3.23381
21.774 3.16718
23.0591 3.03091
24.2651 2.86069
25.3956 2.68442
26.4542 2.52323
27.4435 2.39234
28.3643 2.30177
29.2124 2.22374
30.0971 2.13551
31.0173 2.02552
31.9711 1.87991
32.9604 1.72141
33.878 1.5796
34.726 1.4655
35.5006 1.35527
36.31 1.22388
37.1575 1.09336
38.0447 0.974424
38.968 0.844009
39.9309 0.727159
40.9302 0.599805
41.9693 0.487975
43.0496 0.408933
44.1676 0.339144
45.3186 0.262773
46.5086 0.207159
47.7379 0.19428
49.001 0.199708
};
\addplot [color4, line width=1pt, forget plot]
table {%
0 0
1.38889 -0.0271606
2.81505 -0.0548321
4.27929 -0.0546085
5.78079 -0.053034
7.31986 -0.0815171
8.89485 -0.175478
10.5036 -0.311091
12.149 -0.454757
13.8346 -0.56878
15.5622 -0.61087
17.3267 -0.603188
19.1138 -0.5163
20.8194 -0.313678
22.4299 0.0297115
23.9388 0.467864
25.3551 0.951251
26.6908 1.44057
27.9576 1.90505
29.1792 2.32609
30.3551 2.68247
31.4872 2.95985
32.5797 3.14919
33.6205 3.24269
34.6009 3.23895
35.5117 3.14128
36.3474 2.99174
37.2088 2.79123
38.0955 2.54499
39.0079 2.26004
39.9488 1.94492
40.9215 1.6103
41.9312 1.26883
42.9831 0.935834
44.0827 0.629646
45.2341 0.372121
46.4392 0.189043
47.6954 0.110347
48.993 0.169855
};
\addplot [color5, line width=1pt, forget plot]
table {%
0 0
1.38889 0.0271606
2.70343 0.104201
4.05313 0.206647
5.43599 0.365325
6.85194 0.555578
8.18801 0.789194
9.44704 1.03319
10.6344 1.26128
11.8633 1.47251
13.1371 1.64323
14.4571 1.74595
15.8162 1.80272
17.2129 1.83528
18.6468 1.86877
20.1176 1.93191
21.623 2.05722
23.1616 2.22044
24.7369 2.39082
26.3529 2.5337
28.012 2.60998
29.7116 2.57576
31.4409 2.38263
33.1845 2.04912
34.8346 1.63859
36.4012 1.20202
37.8957 0.780431
39.3278 0.406539
40.8112 0.0823816
42.241 -0.138159
43.6127 -0.237884
44.9129 -0.257704
46.2413 -0.231564
47.6048 -0.127099
48.9752 0.0850637
};
\addplot [color6, line width=1pt, forget plot]
table {%
0 0
1.38889 0.0271606
2.70343 0.104201
3.94005 0.247036
5.09781 0.419159
6.28961 0.612367
7.5111 0.85637
8.76473 1.12671
10.0553 1.40279
11.3881 1.66077
12.7677 1.87302
14.1962 2.00768
15.6651 2.08622
17.1721 2.13429
18.7163 2.18123
20.2972 2.26025
21.9117 2.4086
23.5582 2.60272
25.2409 2.80726
26.9642 2.98262
28.7308 3.08454
30.5396 3.06378
32.378 2.86604
34.2278 2.50832
35.9805 2.05996
37.6472 1.57813
39.2412 1.10922
40.7739 0.69075
42.2522 0.352747
43.6779 0.118676
45.0466 0.00570631
46.4393 -0.0293019
47.7578 -0.0112585
49.0019 0.0272641
};
\addplot [color7, line width=1pt, forget plot]
table {%
0 0
1.38889 0.0271606
2.81505 0.0548322
4.27818 0.111855
5.77615 0.230424
7.30757 0.386246
8.86937 0.610296
10.4613 0.878648
12.0892 1.15806
13.6398 1.45553
15.2141 1.8261
16.8138 2.24618
18.4479 2.6815
20.1271 3.09382
21.8605 3.43988
23.6526 3.67054
25.4906 3.80276
27.2619 3.80394
29.0653 3.63496
30.7744 3.33245
32.4999 2.93137
34.248 2.46951
35.9226 2.01686
37.6424 1.58787
39.4173 1.22786
41.2386 0.916958
42.9892 0.62935
44.6732 0.386418
46.2942 0.223884
47.8487 0.146592
49.3317 0.14337
};
\addplot [color8, line width=1pt, forget plot]
table {%
0 0
1.38889 0.0271606
2.81505 0.0548322
4.27818 0.111855
5.66816 0.167275
7.09453 0.252105
8.44763 0.333123
9.72626 0.434807
11.0382 0.588821
12.3838 0.770474
13.7664 0.956297
15.1899 1.11936
16.6571 1.22879
18.1629 1.30855
19.706 1.38795
21.2851 1.50019
22.8963 1.68248
24.5386 1.91122
26.2168 2.15114
27.9371 2.36283
29.7034 2.50203
31.5144 2.5192
33.3587 2.35968
35.2177 2.04004
36.9825 1.62721
38.6629 1.17862
40.2704 0.741355
41.8321 0.349149
43.3549 0.0341628
44.8394 -0.176337
46.2651 -0.259377
47.6528 -0.201425
48.9866 0.0128232
};
\addplot [color9, line width=1pt, forget plot]
table {%
0 0
1.38889 -0.0271606
2.81505 -0.0548321
4.27929 -0.0546085
5.78079 -0.053034
7.31986 -0.0815171
8.89611 -0.113754
10.3999 -0.115101
11.8302 -0.0583
13.2958 0.0299053
14.7984 0.121778
16.3404 0.185824
17.9197 0.248257
19.5352 0.343791
21.1836 0.511267
22.7467 0.775105
24.331 1.11956
25.834 1.48708
27.2649 1.8401
28.7394 2.17341
30.1551 2.4334
31.6225 2.61294
33.1393 2.67356
34.6924 2.63414
36.274 2.45893
37.8744 2.1681
39.4937 1.79174
41.0306 1.39069
42.4954 1.00433
43.8967 0.663881
45.2396 0.39362
46.5247 0.211653
47.7483 0.130229
48.9982 0.157936
};

\end{axis}

\end{tikzpicture}~
\begin{tikzpicture}

\definecolor{color1}{rgb}{0.280392156862745,0.338158274815817,0.985162233467507}
\definecolor{color0}{rgb}{0,0,0}
\definecolor{color3}{rgb}{0.166666666666667,0.866025403784439,0.866025403784439}
\definecolor{color2}{rgb}{0.0607843137254902,0.636474236147141,0.941089252501372}
\definecolor{color5}{rgb}{0.613725490196078,0.984086337302604,0.641213314833578}
\definecolor{color4}{rgb}{0.386274509803922,0.984086337302604,0.767362681448697}
\definecolor{color7}{rgb}{1,0.636474236147141,0.338158274815817}
\definecolor{color6}{rgb}{0.833333333333333,0.866025403784439,0.5}
\definecolor{color9}{rgb}{1,0,0}
\definecolor{color8}{rgb}{1,0.338158274815818,0.17162567916636}

\begin{axis}[
xlabel={\small{$x$ [m]}}, xlabel near ticks,
ylabel={\small{$y$ [m]}}, ylabel near ticks,
xmin=-3, xmax=53,
ymin=-2.1, ymax=5.6,
width=4.25cm,
height=3cm,
tick align=outside,
tick pos=left,
x grid style={lightgray!92.02614379084967!black},
y grid style={lightgray!92.02614379084967!black}
]
\addplot [black, mark=*, mark size=3, mark options={solid}, line width=1pt, forget plot]
table {%
-10 -1.75
60 -1.75
};
\addplot [black, mark=*, mark size=3, mark options={solid}, line width=1pt, forget plot]
table {%
-10 5.25
60 5.25
};
\addplot [color0, line width=1pt, forget plot]
table {%
0 0
1.38889 0.0271606
2.81396 0.110679
4.27366 0.226018
5.7657 0.404243
7.18014 0.605442
8.62328 0.869298
9.98799 1.14914
11.389 1.43628
12.8325 1.70284
14.3129 1.97476
15.7264 2.20593
17.1769 2.44247
18.5595 2.64028
19.9791 2.84344
21.3301 3.00989
22.7123 3.18096
24.023 3.33559
25.3748 3.46099
26.6587 3.52318
27.8722 3.50807
29.1205 3.44725
30.4017 3.30782
31.6027 3.12866
32.8361 2.92025
34.0973 2.65202
35.2803 2.37345
36.4988 2.08348
37.7465 1.75385
38.9219 1.43862
40.1381 1.13142
41.3972 0.855875
42.6955 0.586937
43.9304 0.372956
45.2108 0.215933
46.5382 0.144919
47.7925 0.139725
49.0355 0.169237
};
\addplot [color1, line width=1pt, forget plot]
table {%
0 0
1.38889 0.0271606
2.81396 0.110679
4.16019 0.270397
5.42597 0.468613
6.61561 0.674729
7.84257 0.883912
9.10232 1.12412
10.2904 1.34864
11.5108 1.60409
12.66 1.84205
13.8507 2.06443
15.0775 2.2969
16.2367 2.49317
17.4303 2.69911
18.6649 2.88733
19.9436 3.03409
21.2611 3.16347
22.6209 3.2449
24.0013 3.24864
25.4172 3.14316
26.75 2.9641
28.1112 2.7308
29.504 2.46562
30.9335 2.194
32.3947 1.88765
33.7838 1.59547
35.2157 1.32434
36.6955 1.10593
38.2162 0.916904
39.6646 0.740594
41.1468 0.531655
42.6657 0.316467
44.2216 0.128075
45.8223 0.00242595
47.4489 -0.0604499
49.0034 -0.0646769
};
\addplot [color2, line width=1pt, forget plot]
table {%
0 0
1.38889 0.0271606
2.81396 0.110679
4.27025 0.283454
5.75371 0.522832
7.15869 0.781976
8.6002 1.04846
10.084 1.29285
11.6049 1.54109
13.0582 1.74923
14.4391 1.94565
15.7441 2.15736
17.086 2.37358
18.4607 2.62275
19.872 2.8785
21.2509 3.10061
22.6295 3.26771
24.0073 3.35664
25.4274 3.34107
26.7693 3.24904
28.1397 3.05247
29.4236 2.79485
30.6258 2.51184
31.8601 2.2044
33.1318 1.89284
34.4344 1.54956
35.667 1.22858
36.9432 0.925601
38.2681 0.666593
39.6342 0.428286
41.047 0.24144
42.3901 0.09457
43.7764 -0.000195183
45.0905 -0.0616567
46.4443 -0.0706264
47.7581 -0.0168672
48.996 0.0664366
};
\addplot [color3, line width=1pt, forget plot]
table {%
0 0
1.38889 0.0271606
2.81396 0.110679
4.27025 0.283454
5.75371 0.522832
7.26751 0.802048
8.81737 1.09103
10.4099 1.35587
11.9308 1.60411
13.3749 1.86891
14.7466 2.12178
16.1603 2.35391
17.5017 2.57354
18.8844 2.77225
20.1945 2.96059
21.4357 3.11432
22.707 3.22416
24.0077 3.2644
25.3311 3.20697
26.5747 3.08496
27.845 2.867
29.1372 2.57486
30.4532 2.22728
31.7979 1.84552
33.1784 1.45372
34.603 1.07948
36.079 0.754571
37.6109 0.515673
39.1876 0.342793
40.6944 0.21534
42.1329 0.152568
43.6089 0.118785
45.0116 0.0882768
46.3418 0.0854506
47.707 0.0813054
48.9991 0.0944744
};
\addplot [color4, line width=1pt, forget plot]
table {%
0 0
1.38889 0.0271606
2.81396 0.110679
4.27025 0.283454
5.6444 0.505194
7.05017 0.759989
8.49251 1.02201
9.96652 1.31966
11.3682 1.60482
12.8124 1.86933
14.2924 2.13895
15.7055 2.36793
17.1539 2.60198
18.5344 2.79745
19.8496 2.93153
21.2002 3.04556
22.5874 3.16356
24.0137 3.25683
25.3703 3.29178
26.764 3.30211
28.0848 3.26031
29.4398 3.14046
30.7111 2.92869
31.8963 2.66671
33.1081 2.35881
34.3505 2.02271
35.6294 1.6789
36.9516 1.35107
38.3229 1.06673
39.7356 0.802499
41.0862 0.604961
42.4767 0.427881
43.8009 0.311698
45.1637 0.216225
46.4568 0.175765
47.7864 0.15621
49.0431 0.160998
};
\addplot [color5, line width=1pt, forget plot]
table {%
0 0
1.38889 0.0271606
2.81396 0.110679
4.27025 0.283454
5.75371 0.522832
7.15869 0.781976
8.49141 1.02834
9.74769 1.28608
11.0409 1.54919
12.2674 1.77383
13.4213 1.98773
14.6161 2.18514
15.8473 2.39189
17.1196 2.58005
18.4361 2.72449
19.7788 2.84817
21.1573 2.97614
22.5744 3.0798
24.022 3.18505
25.3986 3.2579
26.8153 3.27758
28.2689 3.26847
29.7594 3.19939
31.2816 3.03463
32.7146 2.75607
34.0547 2.40787
35.3091 2.0297
36.4864 1.65259
37.7001 1.26672
38.9573 0.893746
40.2648 0.558995
41.6275 0.292105
43.046 0.127483
44.5075 0.0464382
46.006 0.0258067
47.5403 0.0386531
49.0012 0.0533903
};
\addplot [color6, line width=1pt, forget plot]
table {%
0 0
1.38889 0.0271606
2.81396 0.110679
4.27025 0.283454
5.6444 0.505194
7.05017 0.759989
8.49251 1.02201
9.97703 1.2618
11.4987 1.50534
12.9527 1.70898
14.4416 1.91606
15.8543 2.14077
17.1944 2.35453
18.5673 2.60113
19.8679 2.83466
21.2067 3.04811
22.5855 3.21377
24.0102 3.30095
25.4768 3.27376
26.9745 3.15219
28.4961 2.90069
29.9242 2.56556
31.3718 2.16116
32.7393 1.7459
34.1422 1.31924
35.5896 0.909796
37.0898 0.550476
38.6469 0.27946
40.2508 0.0768874
41.7856 -0.0771744
43.2534 -0.163743
44.7376 -0.161024
46.2306 -0.0945805
47.648 3.71411e-05
48.9921 0.0901509
};
\addplot [color7, line width=1pt, forget plot]
table {%
0 0
1.38889 0.0271606
2.81396 0.110679
4.16019 0.270397
5.42597 0.468613
6.61561 0.674729
7.84257 0.883912
9.00132 1.05823
10.1964 1.2419
11.3158 1.43643
12.4728 1.63308
13.6707 1.81268
14.905 2.00106
16.0706 2.15569
17.1629 2.3041
18.2895 2.47966
19.4467 2.70232
20.5285 2.92549
21.6468 3.15138
22.8053 3.36191
23.9999 3.56992
25.2377 3.74906
26.5208 3.87428
27.8485 3.91659
29.2147 3.84312
30.608 3.67377
31.9111 3.39566
33.2276 3.02391
34.4538 2.61314
35.7032 2.15782
36.9745 1.68153
38.2855 1.20435
39.5407 0.78773
40.8184 0.42882
42.1552 0.146094
43.5509 -0.0255426
44.9904 -0.103581
46.3629 -0.0846555
47.6585 -0.00844932
48.9886 0.100923
};
\addplot [color8, line width=1pt, forget plot]
table {%
0 0
1.38889 0.0271606
2.81396 0.110679
4.27024 0.283454
5.6444 0.505193
7.05017 0.759988
8.4925 1.022
9.97702 1.2618
11.4987 1.50534
12.9527 1.70897
14.4436 1.91634
15.8636 2.14221
17.2109 2.35726
18.5674 2.54665
19.9606 2.74178
21.2852 2.90095
22.6466 3.06582
24.0273 3.23407
25.4486 3.38019
26.8022 3.46649
28.0867 3.47526
29.4076 3.38776
30.6444 3.1888
31.7915 2.92401
32.9599 2.59752
34.1526 2.22454
35.3746 1.82258
36.6331 1.41182
37.9358 1.01563
39.2901 0.66125
40.7013 0.3805
42.1698 0.210146
43.6818 0.132069
45.1211 0.122137
46.4863 0.141943
47.7782 0.160992
48.9966 0.199514
};
\addplot [color9, line width=1pt, forget plot]
table {%
0 0
1.38889 0.0271606
2.81396 0.110679
4.27025 0.283454
5.6444 0.505194
7.05017 0.759989
8.49251 1.02201
9.97703 1.2618
11.4987 1.50534
12.9436 1.76567
14.3161 2.01429
15.7306 2.24205
17.1819 2.47507
18.5567 2.72347
19.8593 2.95878
21.1979 3.17363
22.5731 3.39525
23.9769 3.60125
25.4248 3.76394
26.8076 3.8432
28.1179 3.87494
29.3561 3.84163
30.6283 3.72204
31.8195 3.5535
32.9327 3.36622
33.9654 3.14241
35.0203 2.84667
36.0979 2.50068
37.2017 2.11756
38.3371 1.71265
39.5072 1.30514
40.7201 0.91563
41.9856 0.567679
43.3082 0.290171
44.6818 0.117613
46.0893 0.0306633
47.5295 0.00672024
48.9948 0.0785397
};

\end{axis}

\end{tikzpicture}
\caption{Experiment 3. Trajectories for 10 different initial random seeds are displayed for (Left) explicitly accounting for the obstacle points of the dynamic obstacle vehicle vs. (Right) conducting obstacle avoidance via an auxiliary setpoint in the neighboring lane to implicitly account for the dynamic obstacle vehicle. Other quantifications are (Left) $\bar{\tau}_{a}=0.090$s and $v^\text{min}/\bar{v}/v^\text{max}$=28.2/50.6/67.8km/h, and (Right) $\bar{\tau}_{a}=0.085$s and $v^\text{min}/\bar{v}/v^\text{max}$=38.0/49.4/58.6km/h.}
\label{fig:ex4_planarOverlay}
\end{figure}
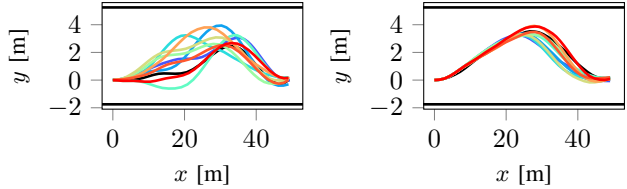

\subsection{Vehicle model \label{subsec_vehMdl}}

For the simulation experiments in the next section a kinematic 4-states-2-controls vehicle model is used:
\begin{equation}
\begin{bmatrix} \dot{x}\\ \dot{y} \\ \dot{\varphi} \\ \dot{v}  \end{bmatrix} = \begin{bmatrix} v \cos(\varphi + \beta)/\cos(\beta) \\ v \sin(\varphi + \beta)/\cos(\beta) \\ v \tan(\delta)/(l_f+l_r) \\ u_v \end{bmatrix},
\end{equation}
with $\beta=\text{atan}(l_r \tan(\delta)/(l_f+l_r))$, states $z=[x,~y,~\varphi,~v]$, and controls $\delta=\delta^\text{max}a[0]$ and $u_v=u_v^\text{min} + \frac{a[1]+1}{2}(u_v^\text{max}-u_v^\text{min})$. All constants are summarized numerically in Table \ref{tab_vehParameters}.

\begin{table}
\vspace{0.3cm}
\centering
\begin{tabular}{|c|c|c|c|c|c|}
\hline
$l_f$ & $l_r$  & $\delta^\text{max}$ & $\dot{\delta}^\text{max}$ & $u_v^\text{min}$ & $u_v^\text{max}$ \\\hline
1.1 & 1.4 & $\frac{40\pi}{180}$ & $\frac{20\pi}{180}$ & $-\frac{100}{3.8*3.6}$ & $\frac{100}{7.4*3.6}$\\\hline
\end{tabular}
\caption{Constants for the vehicle model. All in SI-units. The vehicle chassis has rectangular dimensions $(l_f+0.7+l_r+0.6)\times 2$m.}
\label{tab_vehParameters}
\vspace{-0.4cm}
\end{table}

\setlength\figureheight{6.5cm}
\setlength\figurewidth{6.5cm}
\begin{figure*}
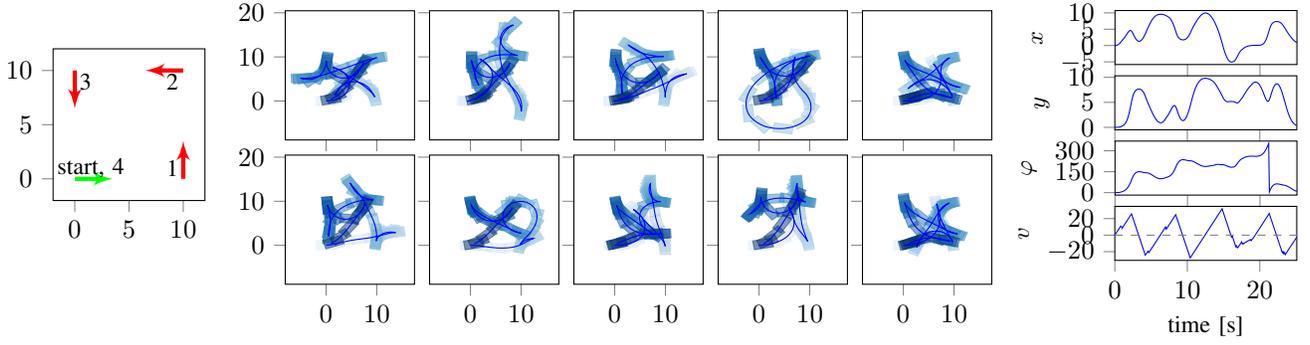

\begin{tabular}{p{0.15\textwidth} p{0.56\textwidth} p{0.12\textwidth}}
\vspace{20pt} 
\begin{tikzpicture}

\begin{axis}[
xmin=-2, xmax=12,
ymin=-2, ymax=12,
width=3.6cm,
height=3.6cm,
tick align=outside,
tick pos=left,
x grid style={lightgray!92.02614379084967!black},
y grid style={lightgray!92.02614379084967!black}
]


\draw[->,color=green!100,>=latex',ultra thick] (axis cs:0,0) -- (axis cs:3.5,0);
\node[color=black] (a) at (axis cs: 1.5,1) {\small{start, 4}};

\draw[->,color=red!100,>=latex',ultra thick] (axis cs:10,0) -- (axis cs:10,3.5);
\node[color=black] (a) at (axis cs: 9,1) {\small{1}};

\draw[->,color=red!100,>=latex',ultra thick] (axis cs:10,10) -- (axis cs:6.5,10);
\node[color=black] (a) at (axis cs: 9,9) {\small{2}};

\draw[->,color=red!100,>=latex',ultra thick] (axis cs:0,10) -- (axis cs:0,6.5);
\node[color=black] (a) at (axis cs: 1,9) {\small{3}};

\end{axis}

\end{tikzpicture} &
\vspace{0pt} \input{Ex5_planarGrid.tex} & 
\vspace{0pt} \input{Ex5_statesSingle0.tex}
\end{tabular}
\vspace{-0.2cm}
\caption{Experiment 4. (Left) Mission formulation: 4 static waypoints (labeled 1-4) are meant to be tracked. (Center) Trajectories for 10 different initial random seeds. (Right) Vehicle states for the grid view at the top-left. Note that in all 10 cases all of the 4 waypoints are successfully tracked, requiring alternating forward and reverse motion and extensive steering. The influence of different random seeds and limited preview in closed-loop is clearly visible.}
\label{fig:ex5_planarGrid}
\vspace{-0.2cm}
\end{figure*}

\section{Numerical Simulation Experiments\label{sec_expts}}

All experiments use the same hyperparameters, see Table \ref{tab_hyperparameters}. All displayed results are \emph{closed-loop} results at sampling rate 0.1s. Controls are throughout initialized as $a_0=[0,~a_0^\text{idle}[1]]$ with $a_0^\text{idle}[1]=-1-2\frac{u_v^\text{min}}{u_v^\text{max}-u_v^\text{min}}$, which implicates zero initial acceleration. All simulation experiments were conducted on an Intel i7-7700K CPU{@}4.20GHz$\times$8 and 1 TitanV-GPU. If plotted, the vehicle chassis is displayed every 0.5s and darker color (blue and red) means later in time. Throughout, standard multilayer perceptrons are used as NN- parametrization. For notation, NN-[5,2,2] denotes an input, hidden, and output layer of 5, 2, and 2 units, respectively.

\begin{table}
\centering
\begin{tabular}{|c|c|c|c|c|c|}
\hline
$T_s$ & $H$  & $N_\text{restarts}$ & $N_\text{iter}^\text{max}$ & $n$ & $N_\text{obstPts}$ \\\hline
0.1 & 200 & 15 & 1 & 20480 & $20$\\\hline
\end{tabular}\vspace{0.15cm}
\begin{tabular}{|c|c|c|c|c|c|c|c|}
\hline
$\epsilon_\xi$ & $\epsilon_\eta$ & $\epsilon_\varphi$ & $\epsilon_v$ & $\Delta \xi$ & $\Delta \eta$ & $\Delta\varphi$ & $\Delta v$ \\\hline
$1.0$ & $0.25$ & $\frac{10\pi}{180}$ & $\frac{5}{3.6}$ & $30$ & $3.5$ & $2\pi$ & $\frac{120}{3.6}$ \\\hline
\end{tabular}
\caption{All experiments use the same hyperparameters. All in SI-units.}
\label{tab_hyperparameters}
\vspace{-0.4cm}
\end{table}

\subsection{Experiment 1: Characteristics of method in view of RRTs}

The purpose of this experiment is to discuss a conceptional issue of RRTs that is not an issue for the proposed method. The experimental setup is $z_0=[0,~0,~0,~0]$, $\zeta^\text{goal}=[0,~0,~\pi,~0]$, no obstacle points and simulation over 10 different random seeds.

Suppose the RRT-algorithm \cite{lavalle1998rapidly} with a tree that is rooted at the origin $z_0$. Suppose now that during the course of RRT-iterations at one point (Step 3 in Alg. of \cite{lavalle1998rapidly}) it is sampled $\zeta^\text{goal}=[0,~0,~\pi,~0]$. Then the next question is what node in the tree to connect the sample to. The original RRT algorithm connects the sample to the ``closest'' node in the tree according to a \emph{distance metric}. This ``select nearest neighbor''-step (Step 4 in Alg. of \cite{lavalle1998rapidly}) is key to any RRT-algorithm because it is responsible for the rapid expansion of the state space. However, at the same time it is also cause of controversy since it is in general not straightforward to decide what distance metric to use for that decision. 
\begin{enumerate}
\item Suppose \emph{Euclidean distance} is used as metric.  Then, the root node $z_0$ would be closest and the 0-input would be requested (Step 5 in Alg. of \cite{lavalle1998rapidly}) for the transition from $z_0$ to the sample according to the Euclidean distance metric. 
Alternatively, when using the Euclidean distance only as metric for the nearest neighbor selection but not for the input selection (Step 5 in Alg. of \cite{lavalle1998rapidly}), then the original problem is recovered and any progress is prohibited by this looping. In both cases the mission is thus impossible to solve.
\item In \cite{kuwata2009real} \emph{Dubins path length} is used as distance metric. A Dubins path implies (i) constant forward speed, and (ii) to be at any time in either a maximal left-turning, straight or maximal right-turning motion. A Dubins path thus assumes it is possible to change instantaneously from maximal left-turn to maximal right-turn at any velocity. In view of RRT, there is consequently a \emph{model mismatch} between the vehicle model used for the distance metric (Dubins car) and the actual vehicle model with (i) steering rate constraints and (ii) the capability to move both forward and reverse to connect any 2 vehicle poses.
\end{enumerate}
An additional comment to RRT$^*$~\cite{karaman2011anytime} is made. The issue of distance metric selection perseveres also for RRT$^*$. This is because it differs from the original RRT only in (i) the different method for parent node selection for a new sample, and (ii) in adding a step for rewiring tree connections based on cost (estimates) accounting for the new sample node.

This discussion is given to contrast simplicity and generality of the proposed method, which (i) does not involve any model mismatch at any stage (it is always worked with full vehicle dynamics and all of its contraints and motion capabilities), (ii) controls both steering and acceleration coupledly, and (iii) permits to sample unconstrainedly in the parameter space of the NN, i.e., $\theta\in(-\infty,\infty)$, which is ideal for exploration, and (iv) the fact that controls $a_{t+h}=\mathcal{X}(s_{t+h},\theta),\forall h=0,\dots,H$, are varying over horizon $H$ even for a small NN-[5,2,2] with only $|\theta|=18$ parameters. Results for the experiment are displayed in Fig. \ref{fig:exrrt_planarGrid}. The influence of different initial random seeds is clearly visible (very different behaviors are obtained). Nevertheless, the mission is solved for all of them. Note that desirable 3-point steering is obtained in 9 out of 10 cases eventhough \emph{no} auxiliary setpoints were set and 3-point-steering was thus not a priori encouraged in any form. Instead, it evolved naturally.


\subsection{Experiment 2: Effect of NN-size\label{subsec_Ex2}}

\begin{table}
\centering
\begin{supertabular}{|l|c|c|c|c|}
\hline
NN-architecture & $|\theta|$ & $|y|^\text{max}$ & $v^\text{min}$/$\bar{v}$/$v^\text{max}$ & $\bar{\tau}_{a}$\\\hline
$[5,2,2]$ & 18 & 0.51 & 33.5/50.4/64.9 & \textbf{0.036} \\
$[5,10,2]$ & 82 & 0.29 & 28.0/46.0/65.0 & 0.047 \\
$[5,10,10,2]$ & 192 & 0.26 & 30.2/50.1/65.1 & \textbf{0.130} \\\hline
$[5,2,2]\dagger$ & 18 & 0.32 & 37.9/51.6/63.5 & 0.036 \\\hline
\end{supertabular}
\caption{Experiment 2. Two computation times are in bold for emphasis. The number of parameters for each NN is denoted by $|\theta|$. $\dagger$: The experimental setup is identical, except that $\zeta^\text{goal}[0]=25$m is used instead of 50m as in the other cases.}
\label{tab_ex2}
\vspace{-0.4cm}
\end{table}

The experimental setup is $z_0=[0,~0,~0,~50/3.6]$, $\zeta^\text{goal}=[50,~0,~0,~50/3.6]$, no obstacle points and simulation over 10 different random seeds. The objective is to analyze influence of NN-size on (i) average computation times $\bar{\tau}_a$ and (ii) on wiggling motion behavior characteristic for sampling based control. The latter is here measured by the maximal lateral overshoot $|y|^\text{max}$ over all 10 simulations. Three NNs are compared: NN-[5,2,2], -[5,10,2] and -[5,10,10,2]. Results are summarized in Fig. \ref{fig_Ex1_522_planarOverlay} and Table \ref{tab_ex2}. The following trade-off is observed: the larger the NN the smaller $|y|^\text{max}$, but the larger also $\bar{\tau}_a$. Maximal lateral overshoot could be reduced from 0.51m to 0.32m (while not compromising on $\bar{\tau}_a$) for the smallest NN-[5,2,2] by just reducing the initial distance of goal setpoint from 50m to 25m, see Table \ref{tab_ex2}. In all subsequent experiments NN-[5,2,2] is employed for its fastest computation times.


\setlength\figureheight{6.5cm}
\setlength\figurewidth{6.5cm}
\begin{figure*}
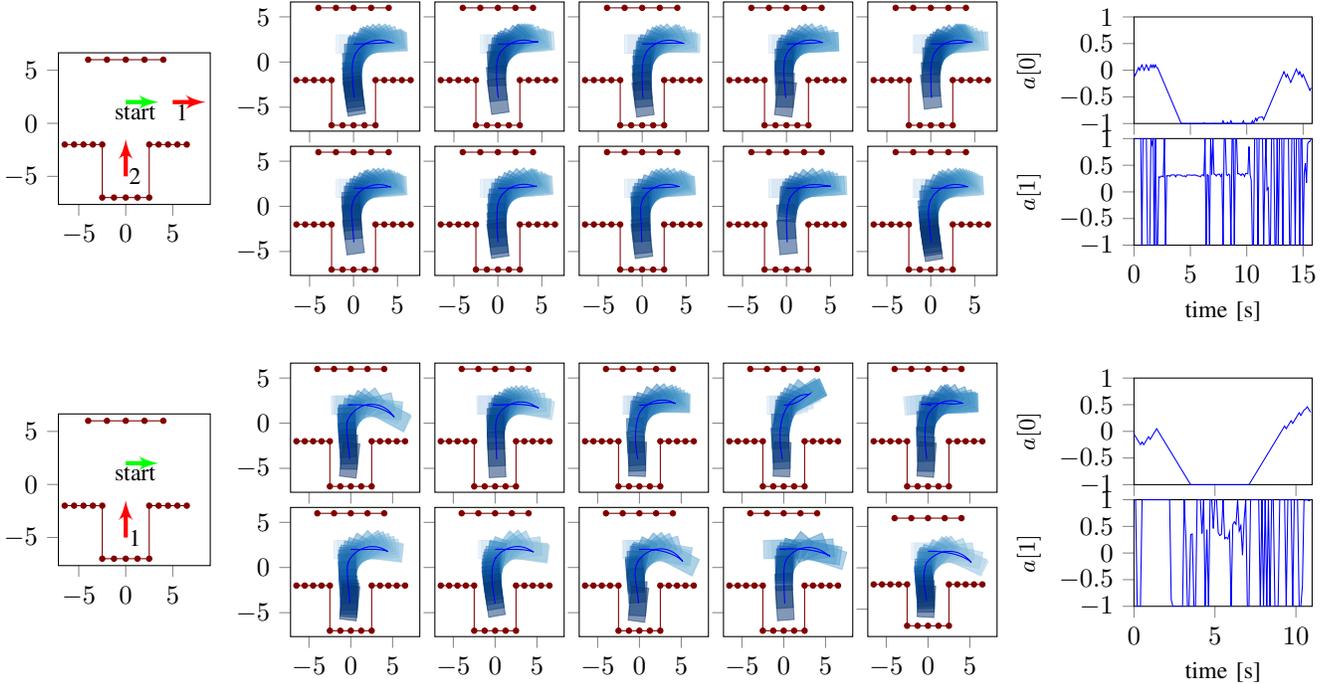

\vspace{0.2cm}
\begin{tabular}{p{0.15\textwidth} p{0.56\textwidth} p{0.12\textwidth}}
\vspace{20pt} 
\begin{tikzpicture}

\begin{axis}[
xmin=-7.16858337272437, xmax=9,
ymin=-7.65, ymax=6.65,
width=3.6cm,
height=3.6cm,
tick align=outside,
tick pos=left,
x grid style={lightgray!92.02614379084967!black},
y grid style={lightgray!92.02614379084967!black}
]


\draw[->,color=green!100,>=latex',ultra thick] (axis cs:0,2) -- (axis cs:3.5,2);
\node[color=black] (a) at (axis cs: 1,1) {\small{start}};

\draw[->,color=red!100,>=latex',ultra thick] (axis cs:5,2) -- (axis cs:8.5,2);
\node[color=black] (a) at (axis cs: 6,1) {\small{1}};

\draw[->,color=red!100,>=latex',ultra thick] (axis cs:0,-5) -- (axis cs:0,-1.5);
\node[color=black] (a) at (axis cs: 1,-5) {\small{2}};

\addplot [red!50.0!black, mark=*, mark size=1, mark options={solid}, forget plot]
table {%
-6.5 -2
-5.5 -2
-4.5 -2
-3.5 -2
-2.5 -2
-2.5 -7
-1.25 -7
0 -7
1.25 -7
2.5 -7
2.5 -2
3.5 -2
4.5 -2
5.5 -2
6.5 -2
};
\addplot [red!50.0!black, mark=*, mark size=1, mark options={solid}, forget plot]
table {%
-4 6
-2 6
0 6
2 6
4 6
};

\end{axis}

\end{tikzpicture} &
\vspace{0pt} \input{Ex9b_planarGrid.tex} & 
\vspace{0pt} 
\begin{tikzpicture}

\begin{groupplot}[group style={group size=1 by 2,vertical sep=0.2cm}]
\nextgroupplot[
ylabel={\small{$a[0]$}}, ylabel near ticks,
xmin=0, xmax=15.8,
ymin=-1, ymax=1,
width=3.95cm,
height=3cm,
tick align=outside,
tick pos=left,
xticklabels={\empty}, xlabel={\empty},
x grid style={lightgray!92.02614379084967!black},
y grid style={lightgray!92.02614379084967!black}
]
\addplot [blue, forget plot]
table {%
0 -0.05
0.1 -0.1
0.2 -0.05
0.3 0
0.4 0.05
0.5 0
0.6 0.05
0.7 0.1
0.8 0.05
0.9 0
1 0.05
1.1 0.1
1.2 0.05
1.3 0
1.4 0.05
1.5 0.1
1.6 0.05
1.7 0.1
1.8 0.05
1.9 0.1
2 0.0528681
2.1 0.0431365
2.2 -0.00686349
2.3 -0.0568635
2.4 -0.106863
2.5 -0.156863
2.6 -0.206863
2.7 -0.256863
2.8 -0.306863
2.9 -0.356863
3 -0.406864
3.1 -0.456864
3.2 -0.506864
3.3 -0.556864
3.4 -0.606864
3.5 -0.656864
3.6 -0.706864
3.7 -0.756864
3.8 -0.806864
3.9 -0.856864
4 -0.906864
4.1 -0.956864
4.2 -1
4.3 -1
4.4 -1
4.5 -1
4.6 -1
4.7 -1
4.8 -1
4.9 -1
5 -1
5.1 -1
5.2 -1
5.3 -1
5.4 -1
5.5 -1
5.6 -1
5.7 -1
5.8 -1
5.9 -1
6 -1
6.1 -1
6.2 -0.988418
6.3 -1
6.4 -1
6.5 -1
6.6 -1
6.7 -1
6.8 -1
6.9 -1
7 -1
7.1 -1
7.2 -0.997409
7.3 -1
7.4 -1
7.5 -1
7.6 -1
7.7 -1
7.8 -1
7.9 -0.95
8 -1
8.1 -1
8.2 -1
8.3 -0.976178
8.4 -1
8.5 -1
8.6 -0.95
8.7 -1
8.8 -1
8.9 -1
9 -1
9.1 -1
9.2 -1
9.3 -1
9.4 -1
9.5 -1
9.6 -1
9.7 -0.950631
9.8 -1
9.9 -1
10 -1
10.1 -1
10.2 -1
10.3 -1
10.4 -1
10.5 -0.95
10.6 -1
10.7 -0.95
10.8 -0.9
10.9 -0.95
11 -0.9
11.1 -0.882234
11.2 -0.8824
11.3 -0.879132
11.4 -0.878226
11.5 -0.928226
11.6 -0.878226
11.7 -0.828226
11.8 -0.778226
11.9 -0.728226
12 -0.678226
12.1 -0.628226
12.2 -0.578225
12.3 -0.528225
12.4 -0.478225
12.5 -0.428225
12.6 -0.378225
12.7 -0.328225
12.8 -0.278225
12.9 -0.228225
13 -0.178225
13.1 -0.128225
13.2 -0.0782254
13.3 -0.0282254
13.4 -0.0782254
13.5 -0.128225
13.6 -0.0782254
13.7 -0.128225
13.8 -0.178225
13.9 -0.228225
14 -0.178225
14.1 -0.128225
14.2 -0.0782254
14.3 -0.0282254
14.4 0.0217746
14.5 -0.0282254
14.6 -0.0782254
14.7 -0.0282254
14.8 -0.0782254
14.9 -0.128225
15 -0.0782254
15.1 -0.128225
15.2 -0.178225
15.3 -0.228225
15.4 -0.278225
15.5 -0.328225
15.6 -0.378225
15.7 -0.328225
};

\nextgroupplot[
xlabel={\small{time [s]}}, xlabel near ticks,
ylabel={\small{$a[1]$}}, ylabel near ticks,
xmin=0, xmax=15.8,
ymin=-1, ymax=1,
width=3.95cm,
height=3cm,
tick align=outside,
tick pos=left,
x grid style={lightgray!92.02614379084967!black},
y grid style={lightgray!92.02614379084967!black}
]
\addplot [blue, forget plot]
table {%
0 1
0.1 1
0.2 1
0.3 1
0.4 1
0.5 1
0.6 1
0.7 -1
0.8 1
0.9 1
1 1
1.1 1
1.2 -1
1.3 -1
1.4 -1
1.5 1
1.6 0.886176
1.7 1
1.8 -1
1.9 1
2 -1
2.1 -1
2.2 0.28248
2.3 0.26596
2.4 0.293799
2.5 0.293646
2.6 0.281806
2.7 1
2.8 -1
2.9 0.303728
3 0.301139
3.1 0.304697
3.2 0.291246
3.3 0.317025
3.4 0.318543
3.5 0.318303
3.6 0.316945
3.7 0.281669
3.8 0.318988
3.9 0.315627
4 0.317775
4.1 0.317315
4.2 0.308244
4.3 0.295346
4.4 0.293778
4.5 0.300989
4.6 0.281567
4.7 0.315874
4.8 0.317679
4.9 0.316231
5 0.314726
5.1 0.310169
5.2 0.310173
5.3 0.320132
5.4 0.312222
5.5 0.313798
5.6 0.301632
5.7 0.297211
5.8 0.293886
5.9 0.282845
6 0.313454
6.1 0.320236
6.2 0.301441
6.3 1
6.4 -1
6.5 0.322059
6.6 0.307349
6.7 -1
6.8 0.999834
6.9 0.327544
7 1
7.1 0.331412
7.2 0.326968
7.3 0.297253
7.4 0.326287
7.5 0.331202
7.6 0.331636
7.7 0.29035
7.8 0.331603
7.9 -1
8 1
8.1 1
8.2 0.319153
8.3 0.305349
8.4 0.333787
8.5 0.314836
8.6 -1
8.7 1
8.8 1
8.9 -1
9 1
9.1 1
9.2 0.338522
9.3 0.340662
9.4 0.327992
9.5 0.306572
9.6 0.338923
9.7 0.330033
9.8 0.29119
9.9 0.345577
10 0.345692
10.1 0.34032
10.2 0.849273
10.3 0.324802
10.4 0.215632
10.5 -1
10.6 -1
10.7 1
10.8 1
10.9 1
11 -1
11.1 1
11.2 1
11.3 1
11.4 1
11.5 -1
11.6 0.221705
11.7 1
11.8 0.0389817
11.9 0.0734985
12 -1
12.1 -1
12.2 1
12.3 1
12.4 1
12.5 -1
12.6 -1
12.7 1
12.8 -1
12.9 1
13 1
13.1 1
13.2 -1
13.3 1
13.4 1
13.5 1
13.6 1
13.7 1
13.8 -1
13.9 -1
14 1
14.1 1
14.2 1
14.3 -1
14.4 1
14.5 1
14.6 1
14.7 -1
14.8 1
14.9 1
15 -1
15.1 0.356134
15.2 0.537
15.3 0.172454
15.4 0.900772
15.5 0.939764
15.6 0.944056
15.7 1
};

\end{groupplot}

\end{tikzpicture}
\end{tabular}
\begin{tabular}{p{0.15\textwidth} p{0.56\textwidth} p{0.12\textwidth}}
\vspace{20pt} 
\begin{tikzpicture}

\begin{axis}[
xmin=-7.16858337272437, xmax=9,
ymin=-7.65, ymax=6.65,
width=3.6cm,
height=3.6cm,
tick align=outside,
tick pos=left,
x grid style={lightgray!92.02614379084967!black},
y grid style={lightgray!92.02614379084967!black}
]


\draw[->,color=green!100,>=latex',ultra thick] (axis cs:0,2) -- (axis cs:3.5,2);
\node[color=black] (a) at (axis cs: 1,1) {\small{start}};


\draw[->,color=red!100,>=latex',ultra thick] (axis cs:0,-5) -- (axis cs:0,-1.5);
\node[color=black] (a) at (axis cs: 1,-5) {\small{1}};

\addplot [red!50.0!black, mark=*, mark size=1, mark options={solid}, forget plot]
table {%
-6.5 -2
-5.5 -2
-4.5 -2
-3.5 -2
-2.5 -2
-2.5 -7
-1.25 -7
0 -7
1.25 -7
2.5 -7
2.5 -2
3.5 -2
4.5 -2
5.5 -2
6.5 -2
};
\addplot [red!50.0!black, mark=*, mark size=1, mark options={solid}, forget plot]
table {%
-4 6
-2 6
0 6
2 6
4 6
};

\end{axis}

\end{tikzpicture} &
\vspace{0pt} \input{Ex9_planarGrid.tex} & 
\vspace{0pt} 
\begin{tikzpicture}

\begin{groupplot}[group style={group size=1 by 2,vertical sep=0.2cm}]
\nextgroupplot[
ylabel={\small{$a[0]$}}, ylabel near ticks,
xmin=0, xmax=11,
ymin=-1, ymax=1,
width=3.95cm,
height=3cm,
tick align=outside,
tick pos=left,
xticklabels={\empty}, xlabel={\empty},
x grid style={lightgray!92.02614379084967!black},
y grid style={lightgray!92.02614379084967!black}
]
\addplot [blue, forget plot]
table {%
0 -0.05
0.1 -0.1
0.2 -0.15
0.3 -0.2
0.4 -0.25
0.5 -0.2
0.6 -0.25
0.7 -0.2
0.8 -0.15
0.9 -0.1
1 -0.15
1.1 -0.1
1.2 -0.05
1.3 -7.45058e-09
1.4 0.05
1.5 -7.45058e-09
1.6 -0.05
1.7 -0.1
1.8 -0.15
1.9 -0.2
2 -0.25
2.1 -0.3
2.2 -0.35
2.3 -0.4
2.4 -0.45
2.5 -0.5
2.6 -0.55
2.7 -0.6
2.8 -0.65
2.9 -0.7
3 -0.75
3.1 -0.8
3.2 -0.85
3.3 -0.9
3.4 -0.95
3.5 -1
3.6 -1
3.7 -1
3.8 -1
3.9 -1
4 -1
4.1 -1
4.2 -1
4.3 -1
4.4 -1
4.5 -1
4.6 -1
4.7 -1
4.8 -1
4.9 -1
5 -1
5.1 -1
5.2 -1
5.3 -1
5.4 -1
5.5 -1
5.6 -1
5.7 -1
5.8 -1
5.9 -1
6 -1
6.1 -1
6.2 -1
6.3 -1
6.4 -1
6.5 -1
6.6 -1
6.7 -1
6.8 -1
6.9 -0.999675
7 -0.99968
7.1 -1
7.2 -0.95
7.3 -0.9
7.4 -0.85
7.5 -0.8
7.6 -0.75
7.7 -0.7
7.8 -0.65
7.9 -0.6
8 -0.55
8.1 -0.5
8.2 -0.45
8.3 -0.4
8.4 -0.35
8.5 -0.3
8.6 -0.25
8.7 -0.2
8.8 -0.15
8.9 -0.0999998
9 -0.0499998
9.1 1.56462e-07
9.2 0.0500002
9.3 0.1
9.4 0.0500002
9.5 0.1
9.6 0.15
9.7 0.2
9.8 0.15
9.9 0.2
10 0.25
10.1 0.3
10.2 0.35
10.3 0.3
10.4 0.35
10.5 0.4
10.6 0.421461
10.7 0.460099
10.8 0.410099
10.9 0.360099
};

\nextgroupplot[
xlabel={\small{time [s]}}, xlabel near ticks,
ylabel={\small{$a[1]$}}, ylabel near ticks,
xmin=0, xmax=11,
ymin=-1, ymax=1,
width=3.95cm,
height=3cm,
tick align=outside,
tick pos=left,
x grid style={lightgray!92.02614379084967!black},
y grid style={lightgray!92.02614379084967!black}
]
\addplot [blue, forget plot]
table {%
0 1
0.1 1
0.2 -1
0.3 -1
0.4 -1
0.5 1
0.6 1
0.7 1
0.8 1
0.9 1
1 1
1.1 1
1.2 1
1.3 1
1.4 1
1.5 1
1.6 1
1.7 1
1.8 1
1.9 1
2 1
2.1 1
2.2 1
2.3 -0.87318
2.4 -1
2.5 -1
2.6 -1
2.7 -1
2.8 -1
2.9 -1
3 -1
3.1 1
3.2 0.314742
3.3 -1
3.4 -1
3.5 0.345539
3.6 0.356594
3.7 -1
3.8 -1
3.9 1
4 1
4.1 1
4.2 0.367209
4.3 1
4.4 -1
4.5 0.436565
4.6 -1
4.7 1
4.8 1
4.9 0.460778
5 1
5.1 1
5.2 0.384426
5.3 0.327757
5.4 0.349023
5.5 1
5.6 0.268662
5.7 0.29401
5.8 0.264722
5.9 0.388812
6 -1
6.1 0.53251
6.2 0.585754
6.3 0.538356
6.4 1
6.5 0.415454
6.6 0.511139
6.7 -0.0382961
6.8 0.616068
6.9 1
7 -1
7.1 -1
7.2 1
7.3 1
7.4 1
7.5 1
7.6 1
7.7 1
7.8 -1
7.9 1
8 -1
8.1 1
8.2 -1
8.3 1
8.4 -1
8.5 1
8.6 1
8.7 -0.992899
8.8 -1
8.9 -1
9 1
9.1 1
9.2 1
9.3 1
9.4 -1
9.5 1
9.6 1
9.7 1
9.8 -1
9.9 1
10 1
10.1 1
10.2 -1
10.3 -1
10.4 -0.618293
10.5 1
10.6 0.992067
10.7 1
10.8 0.979447
10.9 1
};

\end{groupplot}

\end{tikzpicture}
\end{tabular}
\vspace{-0.2cm}
\caption{Experiment 5. TOP ROW: (Left) Mission formulation: 3 static waypoints are tracked. These are ``1'', an auxiliary one with $\zeta^\text{goal}=[0,0,\pi/2,0]$ (not displayed for clarity),  and ``2''. The static obstacle points are indicated by the dots. (Center) Trajectories for 10 different initial random seeds. (Right) Controls output from the NN-[5,2,2] for the grid view at the top-left. Note that the vehicle is operating at its handling limits (both absolute and maximum rate steering constraints are saturated). More quantifications averaged over all 10 random seeds are: $\bar{\tau}_{a}=0.088$s, $v^\text{min}/\bar{v}/v^\text{max}=-17.3/-1.2/16.2$km/h and $P^\text{min}/\bar{P}/P^\text{max}=13.8/14.3/15.2$m. BOTTOM ROW: (Left) Mission formulation: 2 static waypoints are tracked. These are an auxiliary one (not displayed for clarity) with $\zeta^\text{goal}=[0,0,\pi/2,0]$ and ``1''. (Center) Trajectories for 10 different initial random seeds. (Right) Controls output from the NN-[5,2,2] for the grid view at the top-left. Note that the vehicle is operating at its handling limits (both absolute and maximum rate steering constraints are saturated). More quantifications averaged over all 10 random seeds are: $\bar{\tau}_{a}=0.088$s, $v^\text{min}/\bar{v}/v^\text{max}=-16.4/-1.5/19.1$km/h and $P^\text{min}/\bar{P}/P^\text{max}=12.5/15.7/17.8$m.}
\label{fig:ex9b_planarGrid}
\vspace{-0.5cm}
\end{figure*}

\subsection{Experiment 3: Effect of obstacle points\label{subsec_Ex4and6}}


The basic experimental setup is identical to Sect. \ref{subsec_Ex2}, however, (i) a dynamic obstacle moving \emph{towards} the EV on the same lane at 20km/h starting at $x=40$ which must be avoided, and (ii) road-bounds at $y=-1.75$m and 5.25m are added. The objective is (i) to monitor the effect of obstacle points on computation times, and (ii) to compare trajectories when (1) explicitly considering obstacle points of the dynamic obstacle vs. (2) implicitly considering the dynamic obstacle by setting an \emph{auxiliary} setpoint in the neighboring lane and simultaneously discarding obstacle points of the dynamic obstacle. For this experiment, the total number of obstacle points\footnote{Each obstacle point is assumed as a 4D vector with planar location, heading and velocity information, see Sect. \ref{subsec_at}.} considered  is $N_\text{obstPts}=20$ in both scenarios. In the latter, all obstacle points are assigned to define road bounds. Results are displayed in Fig. \ref{fig:ex6_planarGrid} and~ \ref{fig:ex4_planarOverlay}. It is observed that (i) the inclusion of  obstacle points causes an increase from $\bar{\tau}_{a}=0.036$s in the obstacle-free case of Experiment 2 to $\bar{\tau}_{a}=0.090$s here, and (ii) performing dynamic obstacle avoidance via an auxiliary setpoint generates more consistent and smoother motion over different random seeds. Since no velocity constraints were considered (to better observe sampling behavior), there are velocity variations as reported in the caption of Fig. \ref{fig:ex4_planarOverlay}. Note that velocity constraints can easily be enforced by discarding samples (i.e., NN-parametrizations) that violate these.


\subsection{Experiment 4: Effect of randomization}

The experimental setup is $z_0=[0,~0,~0,~0]$, tracking of 4 static setpoints with position and heading as in Fig. \ref{fig:ex5_planarGrid}, no obstacle points and simulation over 10 different random seeds. The objective is to analyze influence of randomization via different initial random seeds and to show capabilities of the method in a complex mission requiring extensive steering and alternating forward and reverse driving. Results are displayed in Fig. \ref{fig:ex5_planarGrid}. For all 10 random seeds all 4 waypoints are successfully tracked. The influence of different random seeds is clearly visible (very different behaviors are obtained). More quantifications are: $\bar{\tau}_{a}=0.035$s, $v^\text{min}/\bar{v}/v^\text{max}=-37.1/1.2/35.0$km/h and $P^\text{min}/\bar{P}/P^\text{max}= 74/90/110$m.


\subsection{Experiment 5: A reverse parking scenario}

The experimental setup is $z_0=[0,~2,~0,~0]$, simulation over 10 different random seeds, and tracking of 3 and 2 waypoints in a reverse parking-like scenario. In both cases, $N_\text{obstPts}=20$ static obstacle points define the parking lot. The objective is to analyze influence of adding a suitable waypoint and to show capabilities of the method in a complex mission requiring extensive steering, alternating forward and reverse driving, and obstacle avoidance. Results are displayed in Fig. \ref{fig:ex9b_planarGrid}. In both cases for 3 and 2 waypoints, the mission is solved for all 10 random seeds. However, consistency of behavior is clearly improved when including a suitably selected third waypoint (``1'' in the top row of Fig. \ref{fig:ex9b_planarGrid}).

\section{More Comments\label{sec_discussion}}

First, a beneficial characteristic of proposed method is that only a low-dimensional set of NN-parameters needs to be sampled at every $T_s$. Because of the NN-approach with, e.g., only $|\theta|=18$ parameters for NN-[5,2,2], complex motion planning over long prediction horizons (e.g., $H=200$) is still feasible. This is since while $\theta$ is held constant, the feature vector $s_{t+h}$ \emph{is} varying. Consequently also controls are  varying with $a_{t+h} = \mathcal{X}(s_{t+h},\theta),~\forall h=0,\dots,H$.

Second, key hyperparameters were identified. Competing interests are on one hand large $H$, $N_\text{restarts}$, $N_\text{obstPts}$ and large NNs, and on the other hand small computation times. Larger $H$ are especially relevant for low-velocity navigation. For $H=200$ and $T_s=0.1$s at $v=5$km/h the spatial look-ahead horizon ist 27.8m. For perspective, in \cite{berntorp2017path} $H=4$ is used (for a 80km/h lane change). Note that $H=4$ for $T_s=0.1$s at $v=5$km/h implies a spatial preview of only 0.56m. Larger $N_\text{restarts}$ were found to improve performance and be much more important than the $N_\text{iter}^\text{max}$-iteration \cite{plessen2019automating}. Therefore, $N_\text{iter}^\text{max}=1$ in Table \ref{tab_hyperparameters}. As illustrated in Table \ref{tab_ex2}, larger NN reduce wiggling, however increase computational time significantly. Likewise, considering many obstacle points significantly increases computation times. This is because at every $h=0,\dots,H$ it has to be iterated over all obstacle points (first their motion extrapolation according to Sect. \ref{subsec_at}, then collision checking). Heuristics to address this are (i) prefiltering of obstacle points deemed most relevant, and (ii) directing more research effort towards the online selection of goal setpoints $\zeta^\text{goal}$. It was found in the experiments of Sect. \ref{subsec_Ex4and6} that motion was much less wiggly when performing dynamic obstacle avoidance by setting an auxiliary waypoint in the neighboring lane instead of explicitly accounting for the obstacle points of the dynamic obstacle.

Third, in general collision checking is considered to be the most expensive computational bottleneck in sampling-based motion planning algorithms \cite{bialkowski2016efficient}. The generality (arbitrary vehicle shapes, possibility to also account for shielded obstacle points, and expansion-possibility to 3D) and simplicity (linear inequality checks) of \eqref{eq_Apsietc_collAvoidIneqs} comes at a cost, namely, the dependency on the resolution accuracy of finite $N_\text{obstPts}$ obstacle points sufficiently characterizing all relevant obstacles.

Fourth, note that (eventhough on powerful hardware) all presented results  were obtained \emph{without} yet any guiding of the sampling distribution by heuristics \cite{urmson2003approaches, ichter2018learning}. For perspective, in \cite[Sect. IV.A]{kuwata2009real} sampling strategies heuristically vary for (i) on a lane, (ii) at an intersection, (iii) in parking lots, (iv) when passing a static obstacle, (v) for 3 different phases of a 3-point turn, and (vi) for reverse driving. Contrary to RRT-based methods, for warm-starting of proposed method it is not decisive \emph{where} to sample spatially, but instead \emph{what motion primitives} to offline pre-encode in the NN. This is because it is then on-top sampled online in the parameter space of the NN. Favorably, pre-encoding of motion primitives can simultaneously provide certificates about base performance. Such certificates may structurally be more valuable than aforementioned heuristics in \cite[Sect. IV.A]{kuwata2009real}, which just \emph{guide} the probabilistic sampling but do not actually provide equivalent certificates about performance.

Fifth, a main limitation of the current implementation of proposed method is that few obstacle points $N_\text{obstPts}$ could be considered in simulations while maintaining a desired long prediction horizon $H$, a large number of restarts $N_\text{restarts}$, and remain within $T_s=0.1$s. The considerations for ongoing work are therefore as follows: Since $H$ must be maintained high to also admit a larger spatial preview at low velocities, two main tuning knobs remain to increase the computation time available for obstacle points collision checks. First, it is hoped that by pre-encoding of motion primitives for warm-starting the sampling efficiency is improved such that $N_\text{restarts}$ can be reduced significantly, ideally, up to $N_\text{restarts}=1$. Second, it is sought to develop, possibly geometric, mappings from mission data and obstacle points to good waypoints $\zeta^\text{goal}$  such that as many as possible obstacle points can be filtered out before feeding to the GPU for collision checking.

Ultimately, as Table \ref{tab_ex2} illustrated, smaller networks result in smaller computation times. The fact that a \emph{small} NN-[5,2,2] with  only $|\theta|=18$ parameters could solve all of above 5 experiments is a very promising sign for future work merging proposed online sampling with offline pre-encoding. This is since in \cite{plessen2018encoding} it was found that tiny NNs are sufficient to offline encode many motion primitives.

%
%
%
%
\section{Conclusion\label{sec_concl}}

A simple method for online sampling in the parameter space of a neural network for GPU-accelerated motion planning of autonomous vehicles was proposed. It is designed for parallelization and therefore well-suited to benefit from continuing advancements in hardware such as GPUs.

There are 2 main avenues for future work. First, preliminary offline encoding of motion primitives in the NN is considered in order to obtain a better warm-start initialization for on-top online sampling, and to obtain  (offline-generated) certificates about base performance. This preliminary offline encoding is expected to accelerate online sampling through better guided randomization. In general, sampling in the parameter space of a NN for control seems particularly promising since NNs are a natural choice for offline pre-encoding of motion primitives. Second, methods for efficient online waypoint selection (guided by traffic rules and geometric consideration) must be devised, possibly also as a function of offline pre-encoded motion primitives to guarantee specific performance.

%
%
\nocite{*}
\bibliographystyle{ieeetr}
\bibliography{myref}
%
%





\end{document}